\documentclass[a4paper,fleqn]{cas-sc}

\usepackage[numbers]{natbib}

\usepackage{float}
\restylefloat{table}
\usepackage{framed,multirow}
\usepackage{amssymb}
\usepackage{latexsym}
\usepackage{url}
\usepackage{xcolor}
\usepackage{amsmath}
\usepackage{graphicx}
\usepackage{hyperref}
\usepackage{comment}
\usepackage{booktabs}
\usepackage{tabularx}
\usepackage{lineno}
\usepackage{longtable}
\usepackage{makecell}
\usepackage{pifont,xcolor}
\usepackage{caption}
\usepackage{subcaption}
\usepackage{supertabular,booktabs}
\usepackage{lscape}
\usepackage{placeins}
\usepackage{xcolor}

\newcolumntype{P}[1]{>{\centering\arraybackslash}p{#1}}

\newcommand*\colourcheck[1]{%
  \expandafter\newcommand\csname #1check\endcsname{\textcolor{#1}{\ding{51}}}%
}

\newcommand*\colourcross[1]{%
  \expandafter\newcommand\csname #1cross\endcsname{\textcolor{#1}{\ding{55}}}%
}

\colourcheck{green}
\colourcross{red}

\def\tsc#1{\csdef{#1}{\textsc{\lowercase{#1}}\xspace}}
\tsc{WGM}
\tsc{QE}

\begin{document}
\let\WriteBookmarks\relax
\def\floatpagepagefraction{1}
\def\textpagefraction{.001}

\shorttitle{A Systematic Review of Few-Shot Learning in Medical Imaging}    

\shortauthors{Pachetti E., Colantonio S.}  

\title [mode = title]{A Systematic Review of Few-Shot Learning in Medical Imaging}  

\author[1,2]{Eva Pachetti}[orcid=0000-0002-1321-9285]
\cormark[1]
\cortext[1]{Corresponding author}
\ead{eva.pachetti@isti.cnr.it}

\credit{Conceptualization, Methodology, Software, Validation, Formal analysis, Investigation, Data Curation, Writing - Original Draft, Writing - Review \& Editing, Visualization}

\affiliation[1]{organization={Institute of Information Science and Technologies “Alessandro Faedo”, National Research Council of Italy (ISTI-CNR)},
            addressline={via Giuseppe Moruzzi 1}, 
            city={Pisa},
            postcode={56124}, 
            state={PI},
            country={Italy}}

\affiliation[2]{organization={Department of Information Engineering, University of Pisa},
            addressline={via Girolamo Caruso 16}, 
            city={Pisa},
            postcode={56122}, 
            state={PI},
            country={Italy}}

\author[1]{Sara Colantonio}[orcid=0000-0003-2022-0804]
\ead{sara.colantonio@isti.cnr.it}
\credit{Conceptualization, Resources,  Writing - Review \& Editing, Supervision, Project administration, Funding acquisition}

\begin{abstract}
The lack of annotated medical images limits the performance of deep learning models, which usually need large-scale labelled datasets. Few-shot learning techniques can reduce data scarcity issues {\color{red} and enhance medical image analysis speed and robustness. This systematic review gives a comprehensive overview of few-shot learning methods for medical image analysis, aiming to establish a standard methodological pipeline for future research reference. With a particular emphasis on the role of meta-learning, we analysed 80 relevant articles published from 2018 to 2023, conducting a risk of bias assessment and extracting relevant information, especially regarding the employed learning techniques. From this, we delineated a comprehensive methodological pipeline shared among all studies. In addition, we performed a statistical analysis of the studies' results concerning the clinical task and the meta-learning method employed while also presenting supplemental information such as imaging modalities and model robustness evaluation techniques. We discussed the findings of our analysis, providing a deep insight into the limitations of the state-of-the-art methods and the most promising approaches. Drawing on our investigation, we yielded recommendations on potential future research directions aiming to bridge the gap between research and clinical practice.}
\end{abstract}


\maketitle


\section{Introduction}
\label{sec:introduction}
\subsection{Rationale}
\noindent The demand for deep learning (DL) models that can generalize well and achieve high performance with limited data is constantly increasing. Few-Shot Learning (FSL) plays a crucial role in addressing this challenge by enabling models to learn from only a few examples, mimicking the way humans naturally learn. In contrast to the typical practice in DL, which involves pre-training models on large datasets and fine-tuning them on specific tasks, FSL allows models to learn effectively with minimal labelled examples. Among the most prominent models that have successfully addressed this limitation is GPT-3 \cite{Brown20}. Unlike traditional models, GPT-3 does not require fine-tuning on specific tasks. Instead, it leverages FSL during inference by being exposed,  for each task, to a few demonstrations for conditioning without updating its parameters \cite{Brown20}. This approach allows GPT-3 to perform various tasks with just a few examples, showcasing the power of FSL in natural language processing.

FSL finds one of its most crucial applications in medical image analysis for several compelling reasons. Firstly, medical datasets are often limited in size due to privacy concerns, high data acquisition costs, and the laborious process of expert annotation. FSL enables models to achieve robust generalization with minimal labelled examples, making it possible to develop effective medical imaging solutions even with scarce data.
Secondly, FSL alleviates the burden of manual annotation by requiring only a few annotated examples for each new task or medical condition. This capability streamlines the annotation process and supports clinicians in their time-consuming tasks.
Moreover, FSL proves particularly valuable for handling rare medical conditions where acquiring sufficient data for traditional DL approaches may be impractical. Leveraging knowledge from more prevalent diseases, FSL empowers models to adapt to new and rare cases with limited examples.
Furthermore, the medical field constantly encounters new diseases, conditions, and imaging modalities. FSL enables medical imaging models to swiftly adapt and learn from a few examples of these novel tasks, facilitating their seamless integration into clinical practice.
Finally, FSL holds potential in personalized medicine, where models must rapidly adapt to analyze images from individual patients. With just a few examples from each patient, FSL allows the model to tailor its analysis based on specific patient characteristics, enhancing the precision of medical diagnoses and treatments.

{\color{red} We believe that a systematic and comprehensive review of the current state-of-the-art (SOTA) in FSL for the medical imaging domain would equip researchers with insights into the most promising FSL methods and help identify areas where applications lack sufficient exploration. This knowledge would ultimately accelerate the integration of FSL approaches into real-world clinical practice.}

\subsection{Objectives}
\label{sec:objectives}
\noindent {\color{red} This systematic review aims to present a comprehensive overview of the SOTA in FSL techniques applied to medical imaging, ultimately providing the reader with a standard methodological pipeline shared across all the examined studies to be used as a reference for further research in this field. While many FSL reviews primarily concentrate on the broader domain of computer vision \cite{Chen19, Rezaei20, Wang20review}, existing reviews addressing FSL for medical imaging \cite{Kotia21, Nayem23} offer only a limited scope within this domain. These reviews include only a modest number of papers, focusing solely on classification and segmentation outcomes, along with a restricted number of clinical applications. Furthermore, they lack statistical analysis of collected results, as well as a discussion on the risk of bias and applicability concerns. Overall, they do not provide an overview of the shared aspects among the investigated techniques. 

With this work, we aim to collect and highlight all the studies that, in the authors' opinion, make substantial and genuine contributions to this domain. Specifically, we focus on the primary applications of DL in medical imaging, namely segmentation, classification, and registration. For each study, we provide readers with risk of bias and applicability analyses, enabling a critical evaluation of their actual scientific impact. The objective is to showcase innovative techniques with proven effectiveness and robustness and leverage them to establish a standardized methodological pipeline applicable to various medical outcomes. This pipeline can serve as a starting point for researchers to develop effective approaches in this area. In addition to our primary objective, this study aims to provide the following insights:

\begin{itemize}
    \item \textbf{Presentation of studies distribution by outcome.} We highlight the distribution of studies across three outcomes: segmentation, classification and registration. 
    \item \textbf{Presentation of studies and results distribution by clinical task.} For each outcome, we analyze the distribution of studies and provide a statistical analysis of their results w.r.t. the clinical task investigated.
    \item \textbf{Presentation of studies and results distribution by meta-learning technique.} Throughout our analysis, we provide special attention to the meta-learning domain as the most utilized approach to tackle FSL issues.  Specifically, for each outcome, we provide a distribution analysis of the meta-learning methods used, as well as a statistical analysis of the studies' results grouped by meta-learning technique.
    \item \textbf{Additional analyses.} In addition to the core analyses, we explore some secondary aspects, such as data usage information, the most commonly used imaging modalities, and the model robustness assessment methods employed. To not affect the manuscript's readability, we will provide such additional information solely in textual form, prioritizing visual representations only for the primary analyses.

\end{itemize}

\noindent In the following, we outline the structure of the manuscript:
\begin{itemize}
    \item We begin with a theoretical introduction to FSL focusing on meta-learning, with a brief overview of Zero-shot Learning (ZSL). Right after, we delve into the most popular meta-learning methods for FSL, highlighting their potential application also to the ZSL domain.
    \item We describe the literature search methods we employed, including the eligibility criteria and the databases analysed. We also illustrate the main features extracted from each study and the synthesis methods applied.
    \item We show the results of our analyses w.r.t. the objectives of our work as well as the analyses performed regarding the risk of bias and applicability concerns.
    \item We discuss our key findings, including highlighting the limitations of current SOTA approaches and exploring potential avenues for future research to guide further exploration in this field.
    \item We summarize the key findings from our work and, based on them, we finally draw conclusions.
\end{itemize}}


\section{Theoretical background}
\label{sec:theory}

\noindent {\color{red} FSL has been receiving significant attention, especially since the advent of meta-learning. This section delves into the theoretical foundations of FSL, focusing on meta-learning methodologies as the most popular approach to tackling FSL problems. By exploring meta-learning, we aim to shed light on its core concepts, techniques, and applications specific to the FSL domain. Furthermore, we offer a brief focus on ZSL and Generalized ZSL (GZSL) domains, highlighting the utilization of several meta-learning methodologies within this realm as well.}

Meta-learning, a.k.a. \emph{learning-to-learn}, is a powerful paradigm that empowers models to rapidly adapt and generalize to new tasks with minimal training examples. Unlike the traditional training scheme where models are trained on data, meta-learning operates on a higher level by training models on \emph{tasks} or \textbf{episodes}. Thus, this form of training is often referred to as \emph{episodic training}.
During training, the meta-learning model is exposed to multiple episodes, each comprising a few examples of a specific task. As a result, the model acquires transferable knowledge and learns to identify common patterns. Consequently, when faced with a new episode during the testing phase, the model can efficiently leverage its acquired meta-knowledge to make accurate predictions, even with limited examples.
The combination of FSL and meta-learning has shown remarkable results, especially where data availability is limited or when handling novel tasks. Below, we provide a more formal formulation of the meta-learning framework, as outlined in \cite{Hospedales21}, {\color{red} which we will leverage to describe the following SOTA methods.} The inner algorithm ($f$) solves the task $i$ by updating the model parameters $\theta$ to $\theta{'}_i$: this phase is called \emph{base learning}. During the \emph{meta-learning} phase, an outer algorithm updates the model parameters $\theta$ across all the tasks according to an outer objective; the updating entity is regulated by a meta-step hyperparameter $\beta$. As pointed out by Hospedales et al. \cite{Hospedales21}, several classic algorithms, such as hyperparameter optimization, can match this definition; however, what actually defines a modern meta-learning algorithm is the definition of an outer objective with the simultaneous optimization of the inner algorithm w.r.t. to this objective. 

\begin{figure*}
    \centering
    \includegraphics[width=\textwidth]{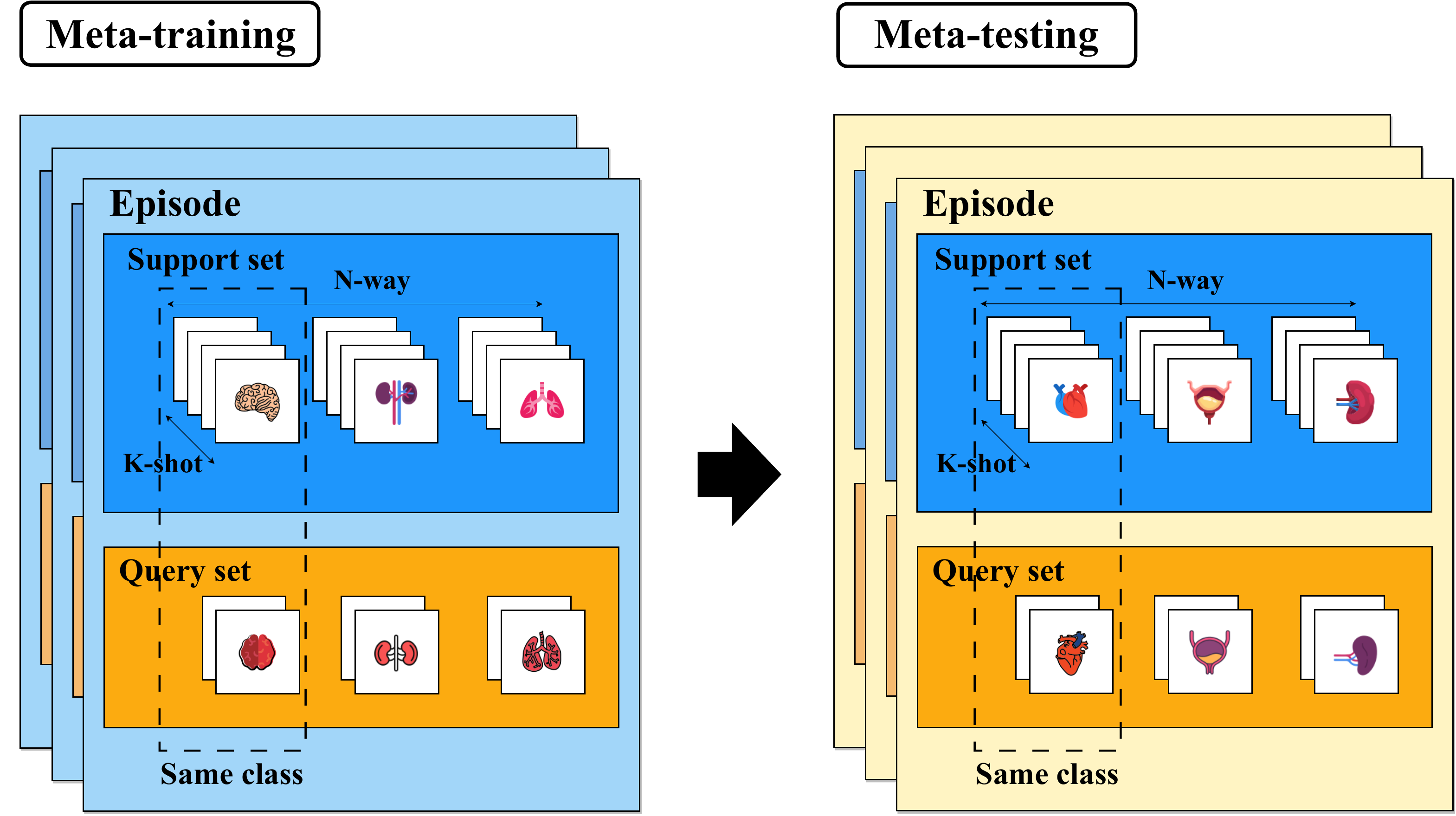}
    \caption{N-way K-shot paradigm representation.}
    \label{fig:N_way_K_shot}
\end{figure*}

A meta-learning training procedure consists of a \emph{meta-training} and a \emph{meta-testing} stage. During meta-training, a set of \emph{source} tasks is sampled from the distribution of the tasks $P(\tau)$. Each source task is composed by a \emph{support} (S = $\{(x_j,y_j)\}^{k}_{j=1}$) and a \emph{query} set (Q = $\{(\hat{x}_j,\hat{y}_j)\}^{k}_{j=1}$), which corresponds to training and validation data in a classical training paradigm, respectively. The goal is to minimize a loss function $\mathcal{L}$ on the query samples conditioned to the support set. During the meta-testing stage, several \emph{target} tasks are sampled as well. In this phase, the base learner is trained on the previously unseen tasks by exploiting the \emph{meta-knowledge} learned during the meta-training phase. To speak about FSL, the number of examples for each class within the support set should be typically less than 10. 
Figure \ref{fig:N_way_K_shot} illustrates the meta-learning training process based on the N-way K-shot paradigm in a generic context where the model's task involves classifying medical images according to the depicted organ.

{\color{red} In scenarios with no shared classes between training and test sets or, more generally, where test samples can be from both training (seen) and unseen classes, we talk about ZSL \cite{Xian17} and GZSL \cite{Pourpanah23}, respectively. For addressing ZSL/GZSL, the most popular approach is to transfer knowledge from seen classes to unseen classes by sharing attributes \cite{Sun21}. In this sense, ZSL/GZSL is often tackled using different methods w.r.t. FSL, like learning intermediate attribute classifiers \cite{Rohrbach11,Lampert13} or learning a mixture of seen class proportions \cite{Zhang15,Changpinyo16} or adopting a direct approach as learning compatibility functions \cite{Palatucci09,Akata15}. Given that a comprehensive discussion of the commonly used methods for addressing ZSL is beyond the scope of this paper, we highlight, in the description of meta-learning for FSL approaches, the ones that have also found utility for ZSL/GZSL.}

\begin{figure*}
    \centering
    \includegraphics[width=12cm]
    {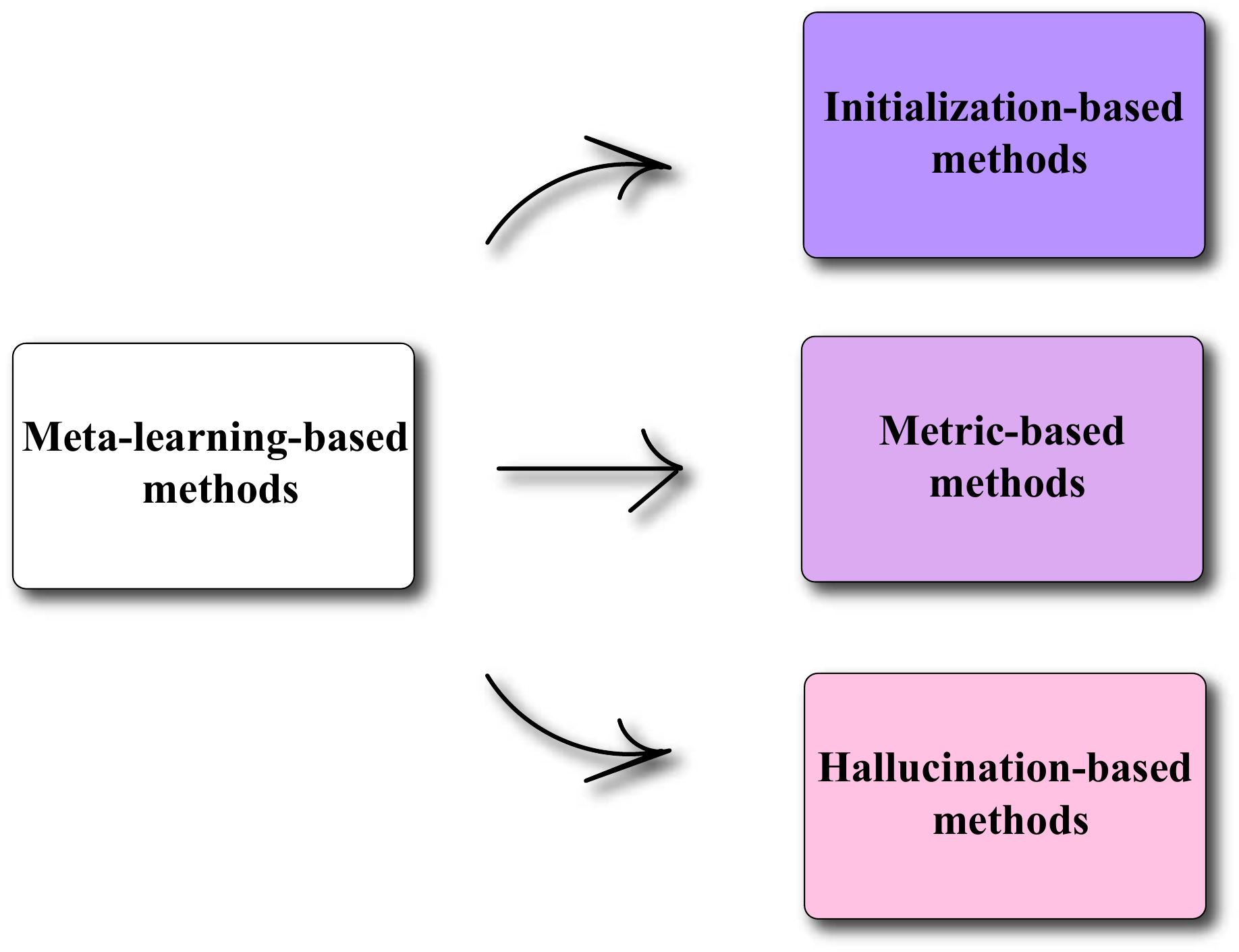}
    \caption{Meta-learning methods taxonomy.}
    \label{fig:meta_l_taxonomy}
\end{figure*}

{\color{red} Since meta-knowledge can manifest in various forms, such as initial parameters, optimization strategy, and learning algorithm \cite{Hospedales21}, we adopt the taxonomy proposed by \cite{Chen19} to categorize meta-learning algorithms for FSL into three categories: \emph{Initialization-based}, \emph{Metric learning-based}, and \emph{Hallucination-based} methods. A representation of the followed taxonomy is provided Figure \ref{fig:meta_l_taxonomy}.} In the subsequent paragraphs, we provide an overview of the most renowned algorithms developed within each category.

\subsection{Initialization-based methods}

\noindent Initialization-based methods refer to a class of approaches that focus on learning effective initializations for model parameters, i.e. \emph{learning to initialize}. The model learns to adjust its parameters or initialization to better adapt to each task during the meta-training phase. The goal is to find parameter initializations that can be readily fine-tuned with only a few examples from a new episode, facilitating rapid generalization. The following are some of the most relevant SOTA algorithms that belong to the category of initialization-based methods in meta-learning.

\subsubsection{Model-Agnostic Meta-Learning}

\noindent In their paper, Finn et al. \cite{Finn17} present Model-Agnostic Meta-Learning (MAML), a meta-learning framework applicable to any model trained with gradient descent. The objective of MAML is to enable the model $f_{\theta}$ to adapt quickly to new tasks $\tau_i$ by finding the model parameters most sensitive to changes in the episode. In particular, the model's parameters are updated to $\theta_i^{'}$ for a new task $\tau_i$ as follows:
\begin{equation} \label{eq:1} \theta_i^{'} = \theta - \alpha \nabla_\theta \mathcal{L}{\tau_i} (f\theta) \end{equation}
where $\alpha$ is the step size of the gradient descent and $\mathcal{L}$ the loss function. The overall meta-objective is to minimize the loss across all tasks $P(\tau)$:
\begin{equation} \label{eq:2} \min_{\theta} \sum_{\tau_i \sim P(\tau_i)} \mathcal{L}{\tau_i}(f{\theta_i^{'}}) \end{equation}
The model parameters are updated through stochastic gradient descent (SGD) as follows:
\begin{equation} \label{eq:3} \theta \leftarrow \theta - \beta\nabla_\theta \sum_{\tau_i \sim P(\tau_i)}\mathcal{L}{\tau_i}(f{\theta_i^{'}}) \end{equation}

Since computing gradients for both task and meta objectives can be computationally expensive, the authors also explored a first-order approximation (FOMAML) that omits the second derivatives. Surprisingly, their results showed that FOMAML performed almost as well as the original MAML. A possible explanation for this observation is that certain ReLU neural networks are nearly linear locally, causing the second derivatives to be close to zero in practice.

{\color{red} In \cite{Verma20}, the authors proposed exploiting MAML to perform GZSL. Specifically, they use the meta-learning framework to train a generative adversarial network conditioned on class attributes that can generate novel class samples. A key difference with classical MAML is that, for each task, the classes between the training and validation phases are disjoint.}

\subsubsection{Reptile} \noindent In their work Nichol, Achiam and Schulman \cite{Nichol18} propose a variant of FOMAML called Reptile. Similar to MAML and FOMAML, Reptile updates the global parameters to create task-specific parameters. However, instead of following Equation \ref{eq:3}, Reptile uses the following update rule for $N$ tasks:

\begin{equation}
    \begin{gathered}
    \theta \leftarrow \theta + \beta \frac{1}{N} \sum_{i=1}^{N} (\theta_i^{'} - \theta)
\end{gathered}
\end{equation}


\noindent Here, the difference $(\theta_i^{'} - \theta)$, instead of being updated towards $\theta$, is treated as a gradient and can be utilized with an adaptive algorithm like Adam for the final update. This update rule is computationally more efficient compared to the complex second-order differentiation used in MAML. This efficiency makes Reptile easier to implement and can lead to faster training times.

\subsubsection{Optimization as Long Short-Term Memory network cell update} \noindent In their work, Ravi and Larochelle \cite{Ravi16} propose a meta-learning approach based on Long Short-Term Memory (LSTM) networks, aiming to learn an optimization algorithm for training another model in an FSL manner. The main idea stems from the observation that the parameter updating law in a generic gradient descent network is similar to the update equation of the cell state in an LSTM \cite{Hochreiter97}:

\begin{align}
    c_t = f_t \odot c_{t-1} + i_t \odot \widetilde{c}_t
\end{align}

\noindent where $f_t = 1$, $c_{t-1} = f_{\theta}$, $i_t = \alpha$, and $\widetilde{c}_t = - \nabla{\theta} \mathcal{L}$. Exploiting this relationship, the learning rate can be formulated as a function of the current parameter value $\theta$, the current gradient $\nabla_{\theta} \mathcal{L}$, the current loss $\mathcal{L}$, and the previous learning rate $\alpha_{t-1}$. By doing so, the meta-learner can effectively control the learning rate value, enabling the model to learn quickly. During training, while iterating on the episode's training set, the LSTM meta-learner receives the values ($\nabla_{\theta} \mathcal{L}{\tau_i}, \mathcal{L}{\tau_i}$) from the model for each task $\tau_i$. Subsequently, it generates the updated parameters $\theta^{'}_{i}$ as its output. This process is repeated for a predefined number of steps, and at the end of these steps, the model's parameters are evaluated on the test set to compute the loss, which is then used for training the meta-learner.

\subsubsection{Optimization with Markov decision process and Reinforcement Learning} \noindent In their paper, Li and Malik \cite{Li16} propose a novel approach to learning an optimization algorithm using guided policy search through reinforcement learning in the form of a Markov decision process (MDP) \cite{Bellman57}. The goal is to learn an optimization algorithm, represented by a \emph{policy} $\pi$, that can efficiently update the current location in an iterative optimization process. The optimization algorithm under consideration performs updates to the current location using a step vector computed by a generic function $\pi$ of the objective function, the current location, and past locations. Each value of $\pi$ corresponds to a different optimization algorithm, so by learning $\pi$, one can effectively learn multiple optimization algorithms. However, learning a generic function $\pi$ is challenging, so the authors restrict the dependence of $\pi$ to the objective values and gradients evaluated at the present and past locations. Consequently, $\pi$ can be modelled as a function that takes the objective values and gradients along the optimizer's trajectory and outputs the next step vector for the optimization.

The authors observe that executing an optimization algorithm can be seen as executing a policy in an MDP, where the current location serves as the state, the step vector as the action, and the transition probability is similar to the location update formula ($x^{(i)} \leftarrow x^{(i-1)} + \Delta x$). The implemented policy corresponds to the choice of $\pi$ used by the optimization algorithm. By searching over policies, they effectively explore a range of possible first-order optimization algorithms. To learn the policy $\pi$, they use reinforcement learning, with the speed of convergence serving as the cost function (policies that lead to slow convergence are penalized). Since $\pi$ could be stochastic in general, the authors use a neural network to parameterize the mean of $\pi$. The current state in the MDP corresponds to the parameters of the neural network, and the system updates these parameters (takes an action from the policy) and receives a reward based on how the loss function changes. 

\subsubsection{Memory-augmented Neural Networks} 
\begin{figure*}
    \centering
    \includegraphics[width=\textwidth]
    {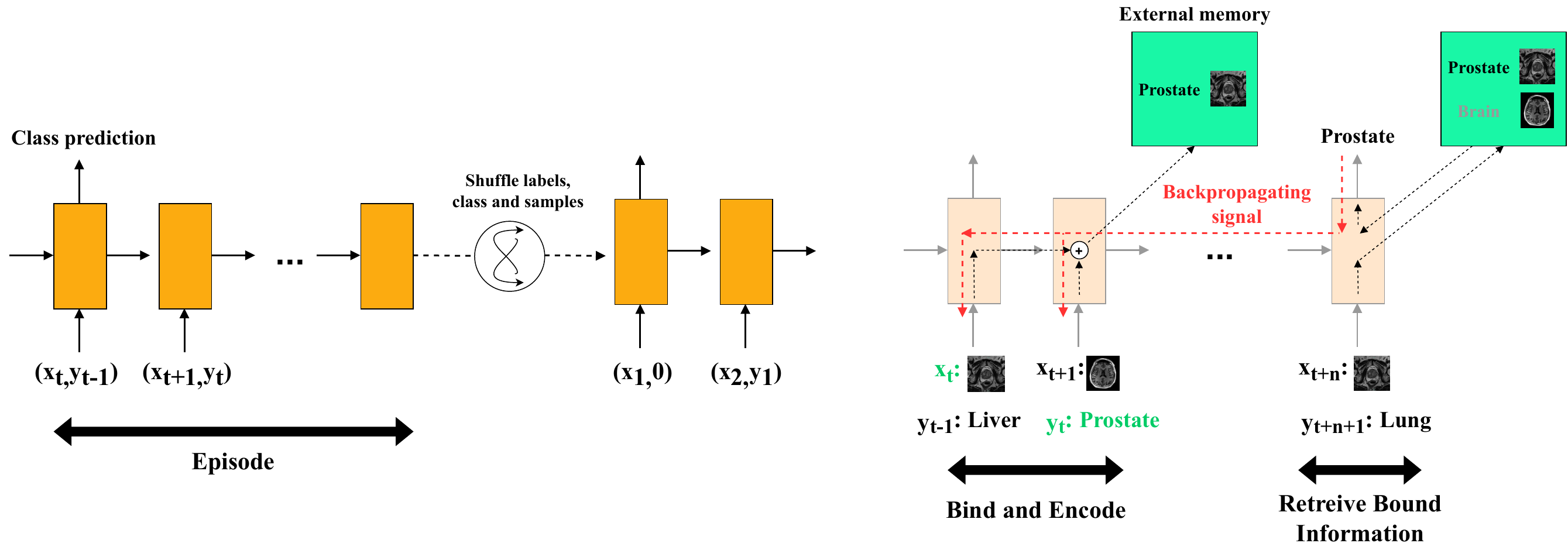}
    \caption{Illustration of meta-learning with MANN approach.}
    \label{fig:mann}
\end{figure*}

\noindent In their paper, Santoro et al. \cite{Santoro16} propose a solution to the FSL task using a differentiable version of Memory-augmented Neural Networks (MANNs) known as Neural Turing Machines (NTMs) \cite{Graves14}.
An NTM consists of a controller, which can be a feed-forward network or a Long Short-Term Memory (LSTM) network, that interacts with an external memory module through reading and writing heads. The NTM's memory reading and writing operations are fast, making it suitable for meta-learning and few-shot predictions.
{\color{red} In the proposed approach, the authors feed} the model with an input while its label is provided a one-time step later. Specifically, at time step $t$, the model receives the input $x_t$ and the label $y_{t-1}$. This approach prevents the model from simply learning to map the label to the output. To further ensure this, inputs and their corresponding labels are shuffled in each episode so that the model cannot learn the input sequence directly.
{\color{red} During the training process, an external memory is utilized to store the input and its bounded label, which are always one time step apart. When an already-seen input shows up, the corresponding label is retrieved from the external memory, providing the input classification. The retrieval process is performed using a key $k_t$ associated with the input $x_t$, produced by the controller and stored in a memory matrix $M_t$. Specifically, the cosine similarity between the key $k_t$ and the memory matrix $M_t$ contents is produced. The error signals from the prediction step are backpropagated to promote this binding strategy. We illustrate the described approach in Figure \ref{fig:mann}.}

\subsection{Metric learning-based methods}

\noindent The metric-learning-based category comprises all the algorithms that enable the model to \emph{learn to compare}. The main idea is to train the model to understand the similarity between images, allowing it to classify a new instance based on its distance w.r.t the seen categories. Below, we report some of the most relevant SOTA metric-learning-based algorithms. 

\subsubsection{Siamese Neural Networks} \noindent Bromley et al. \cite{Bromley93} first introduced Siamese Neural Networks for signature verification. In 2015, they were newly proposed by Koch, Zemel and Salakhutdinov \cite{Koch15}, where they exploited Convolutional Neural Networks (CNNs) to perform one-shot image classification. A Siamese Network consists of two identical networks accepting different inputs and having bound weights to ensure that similar images are mapped close in the feature space. As the network undergoes training, it learns to differentiate between pairs of images that belong to the same class and those that belong to different classes. In the inference phase, a test image is compared with one image per novel class and a similarity score is computed. The network then assigns the highest probability to the pair with the highest score. Because the model is trained on an extensive set of training classes, it becomes proficient at general data discrimination during the training process.

\subsubsection{Triplet Networks} \noindent Triplet Networks, introduced by Hoffer et al. \cite{Hoffer15}, were inspired by Siamese Networks and share the same architectural criterion. Here, the model is composed of three identical networks, having shared parameters, which are trained by triplets composed of anchor, positive and negative samples (positive examples belong to the same class as the anchor, while negative belongs to a different class). The network outputs the $L_2$ distances between the anchor and the positive and negative examples. The objective is to classify which between the positive and negative examples belongs to the same class as the anchor.
During inference time, the model is fed with two inputs and assesses whether they belong to the same class by applying a threshold to the distance in the embedding space.

{\color{red} Triplet networks have also been employed in the ZSL and GZSL domains. Specifically, in \cite{Ji20}, the authors propose a Dual-Triplet Network (DTNet) that employs two triplet networks for learning visual-semantic mapping: one focuses on negative attribute
features, and the other on the negative visual features. Zero-shot classification is performed with a nearest-neighbour approach on the attribute feature, searching for the closest to the test sample.}

\subsubsection{Matching Networks} \noindent Matching Networks proposed by Vinyals et al. \cite{Vinyals16}, differently from Siamese and Triple Networks, can work in a multi-class way instead of in a pair-wise one. Matching Networks aim to map a support set to a classifier, which, given a query example, can produce a probability distribution of the output according to the following equation:

\begin{equation}
    P(\hat{y_j}|\hat{x_j})=\sum_{j=1}^k a(\hat{x_j},x_j)y_j
\end{equation}

\noindent where $a$ acts as an attention mechanism. In the simplest implementation, $a$ consists of computing a softmax over the cosine distance. At each iteration, a training episode is constructed, composed of a support and a query set. Based on the support set, the network provides the query label and the error is minimized.

\subsubsection{Prototypical Networks} \noindent Prototypical Network, proposed by Snell, Swersky, and Zemel \cite{Snell17}, compute a representation or \textit{prototype} of each class using an embedding function with trainable parameters. Given a class $c$, the prototypes are computed by averaging the embeddings of the support samples belonging to each class:

\begin{equation}
    p_c = \frac{1}{|S_c|}\sum_{(x_j,y_j) \in S_c} f_{\theta}(x_j)
\end{equation}

\noindent Given a generic distance function $d$, the prototypical network provides an output distribution based on the distance between the query embeddings and the prototypes of each class:

\begin{equation}
    P(\hat{y_j}=c|\hat{x}_j) = \frac{exp(-d(f_{\theta}(\hat{x}_j),p_c))}{\sum_{c'}exp(-d(f_{\theta}(\hat{x}_j),p_{c'}))}
\end{equation}

\noindent As for Matching Networks, training episodes are built by sampling a set of classes from the training set and choosing two groups of examples for each class as the support and query set, respectively. While in the original paper on Matching Networks, cosine distance was used as a distance function, here, the authors employ the negative squared Euclidean distance (greater distances provide smaller values). As pointed out by the authors, while prototypical networks differ from matching networks in a few-shot scenario, One-Shot Learning (OSL) makes them equivalent.

It is also possible to use this architecture for ZSL. Here, instead of having training points, we have a class meta-data vector for each class, which can be already known or learned, for example, from raw text \cite{Elhoseiny13}. Here, the prototype becomes an embedding of the meta-data vector. 

\subsubsection{Relation Networks} \noindent Relation Networks were introduced by Santoro et al. in their paper \cite{Santoro17}, and they were initially employed in the FSL and ZSL domains in \cite{Sung18}. In contrast to Matching and Prototypical Networks, which use predefined distance functions, a relation network is trained end-to-end, including the metric to compare support and query embeddings. This part of the network is called \textit{relation module}. In a one-shot setting, embeddings from support and query samples are first produced and concatenated in depth through an operator $Z(\cdot,\cdot)$. Concatenated embeddings are provided to the relation module $g_{\phi}$, which outputs a scalar representing the similarity between the support and query embeddings:

\begin{equation}
    r = g_{\phi}(Z(f_{\theta}(x_j),f_{\theta}(\hat{x}_j)))
\end{equation}

\noindent For a generic FSL, the class feature map is calculated by summing all embedding module outputs from each sample in the training set. The class-level feature map is then combined with the query image feature map as in the one-shot scenario.

Relation Networks can be employed in a ZSL as well. In this case, a semantic class embedding vector is provided for each class. Since support and query vectors belong to different modalities (attributes and images, respectively), two embedding modules are employed. The relation module instead works as before.

\subsection{Hallucination-based methods}

\noindent The hallucination-based methods directly address the scarcity of data by \emph{learning to augment}. These methods focus on generating additional data to overcome the limitations of the available dataset. In the following, we describe in detail the most prominent hallucination-based methods.

\subsubsection{Hallucinating with Intra-class Analogies} \noindent Harihan and Girshick \cite{Hariharan17} propose to exploit intra-class analogies to augment the dataset when few examples are available. Their framework employs a learner, two training and a testing phase. In the first training phase, known as \emph{representation learning} phase, the learner is fed with several base classes ($C_{base}$), for which a lot of examples are available for each class. The learner uses these data to set the parameters of its feature extractor. During the second phase (\emph{low-shot} phase), the learner needs to distinguish a set of classes, both base and novel ones. For the novel classes, the learner has access only to a few examples, while for the base classes, it has access to the same dataset used for learning the feature extractor. During the test phase, the model predicts labels from both classes. For the categories with few examples, the idea is to hallucinate additional data using the many examples seen for the base classes to improve the model's performance. The goal is to learn a transformation that maps two images belonging to the same base class (e.g., bird on grass and bird on the sky) and apply this transformation to a novel class image. 
To achieve this, a function $G$ is trained that takes the concatenated feature vectors of three examples and outputs a "hallucinated" feature vector. As $G$, they exploited an MLP with three fully connected layers.

\subsubsection{Classificator and Hallucinator End-to-End Model} \noindent Wang et al. \cite{Wang18} further deepened the previously described method by combining a generator of "hallucinated" examples, with a meta-learning framework, by optimizing the two models jointly. 
The "hallucinator" $G$ takes as input an example $x$, a noise vector $z$ and produces a hallucinated example as the output according to the hallucinator parameters $\theta_G$. During meta-testing, several hallucinated examples are computed by sampling from the initial training set $S_{train}$, producing a new training set $S_{train}^{G}$. The final training set $S_{train}^{aug}$ is obtained by combining the two datasets. This dataset is then used to train the classification algorithm. During the meta-training phase, the hallucinator is trained jointly with the classification algorithm, exploiting a meta-learning paradigm. From the set of all classes, $m$ classes are sampled, specifically $n$ examples for one. The generator $G$ is exploited to produce additional $n$ augmented examples to add to the training set. This new dataset is employed to train the classification algorithm. This training process is agnostic w.r.t. specific meta-learning algorithm used.

After categorizing and describing the main meta-learning methods for FSL in the literature, the following chapter outlines the methods used for searching, selecting, and analyzing SOTA works in the field of FSL for medical image analysis.


\section{Methods}
\label{sec:methods}

\subsection{Study Design}

\noindent We conducted a systematic review in accordance with the “Preferred reporting items for systematic reviews and meta-analyses” (PRISMA) 2020 checklist \cite{Page21}. {\color{red} The primary objective of this review is to establish a comprehensive methodological pipeline shared across all examined studies, serving as a reference for future research in this domain. Additionally, it aims to analyze the distribution of studies across three key outcomes: segmentation, classification, and registration, as well as w.r.t. the clinical tasks investigated and the meta-learning methods employed. Furthermore, it seeks to conduct a statistical analysis of the performance of the studies and provide additional insights, such as the imaging modality analyzed.}

\subsection{Eligilibity criteria}

\noindent We established the inclusion criteria for paper selection based on three primary aspects:
\begin{itemize}
    \item \textbf{Implementation of FSL techniques}: We selected papers that claimed to implement FSL in their work.
    \item \textbf{Application in medical imaging domain}: We considered papers that performed at least one experiment applied to the medical imaging domain. 
    \item \textbf{Low data usage in training}: We included only papers that demonstrated using a small amount of data during training. In particular, we considered all the studies that employed a maximum of 20 training examples per class.
    
\end{itemize}
 
\noindent In addition, during the selection process, we excluded abstracts, non-peer-reviewed papers, papers written in languages other than English, and papers deemed to have significant theoretical errors. Furthermore, we did not include papers dealing with few-shot domain adaptation methods (FSDA), as \cite{Gu22,Keaton23,Li21}. FSDA, as highlighted by Li et al. \cite{Li21}, FSL focuses on adapting pre-trained models to perform well on novel tasks with limited training examples, whereas FSDA involves adapting models across different domains. Therefore, we considered FSDA papers outside the scope of this systematic review.
By applying these inclusion and exclusion criteria, we aimed to ensure the selection of relevant and high-quality papers that specifically addressed the application of FSL techniques in medical imaging with limited training data.

\subsection{Information sources}
 
\noindent We searched for papers using the following databases: 
\begin{itemize}
    \item Web of science
    \item Scopus
    \item IEEE Xplore
    \item ACM Digital Library
\end{itemize}

\noindent To ensure comprehensive coverage and include recent studies in our analysis, we performed a two-step search, the first on September 7, 2022, and the second on January 25, 2023.   
In cases where we didn't have full access to the papers, we took advantage of the Network Inter-Library Document Exchange (NILDE) platform, a web-based Document Delivery service through which we requested access to the missing PDF files, enabling us to obtain the complete papers for inclusion in our review. 

\subsection{Research strategies}

\noindent For each of the mentioned databases, we listed the queries used in the study search in Table \ref{tab:queries}.

\begin{table*}[t]
    \centering
    \begin{tabularx}{\textwidth}{CX}
    \toprule
    \textbf{Database} & \textbf{Query}\\
    \toprule
     Web of Science & (TS=("few-shot") OR TS=("low-shot") OR TS=("one-shot") OR TS=("zero-shot"))  AND (TS=("medical imag*")) AND (TS=("classif*") OR TS=("segment*") OR TS=("regist*")) \\
     \hline
     Scopus & TITLE-ABS-KEY ( {few-shot} )  OR  TITLE-ABS-KEY ( {low-shot} )  OR  TITLE-ABS-KEY ( {one-shot} )  OR  TITLE-ABS-KEY ( {zero-shot} )  AND  TITLE-ABS-KEY ( {medical imaging} )  OR  TITLE-ABS-KEY ( {medical image} )  OR  TITLE-ABS-KEY ( {medical images} )  AND  TITLE-ABS-KEY ( classif* )  OR  TITLE-ABS-KEY ( segment* )  OR  TITLE-ABS-KEY ( regist* )\\
     \hline
     IEEE Xplore & ((("Abstract":"few-shot" OR "Abstract":"low-shot" OR "Abstract":"one-shot" OR "Abstract":"zero-shot") AND "Abstract":"medical imag*" AND ("Abstract":classification OR "Abstract":segmentation OR "Abstract":registration) OR ("Document Title":"few-shot" OR "Document Title":"low-shot" OR "Document Title":"one-shot" OR "Document Title":"zero-shot") AND "Document Title":"medical imag*" AND ("Document Title":classification OR "Document Title":segmentation OR "Document Title":registration) OR ("Author Keywords":"few-shot" OR "Author Keywords"":"low-shot" OR "Author Keywords"":"one-shot" OR "Author Keywords"":"zero-shot") AND "Author Keywords"":"medical imag*" AND ("Author Keywords"":classif* OR "Author Keywords"":segment* OR "Author Keywords"":regist*)) )\\
     \hline
     ACM Digital Library & "(Abstract:(""few-shot"" OR ""low-shot"" OR ""one-shot"" OR ""zero-shot"") OR  Keyword:(""few-shot"" OR ""low-shot"" OR ""one-shot"" OR ""zero-shot"") OR Title:(""few-shot"" OR ""low-shot"" OR ""one-shot"" OR ""zero-shot"")) AND (Abstract:(""medical imaging"" OR ""medical images"" OR ""medical image"") OR Title:(""medical imaging"" OR ""medical images"" OR ""medical image"") OR Keyword:(""medical imaging"" OR ""medical images"" OR ""medical image"")) AND (Title:(classif* OR segment* OR regist*) OR Abstract:(classif* OR segment* OR regist*) OR Keyword:(classif* OR segment* OR regist*))"\\
     \bottomrule
    \end{tabularx}
    \caption{Research queries employed for each database.}
    \label{tab:queries}
\end{table*}

\subsection{Selection process}

\noindent During the review process, a single reviewer examined each record, including titles, abstracts, and any accompanying reports obtained during the research. No machine learning algorithms were employed to aid in eliminating records or to streamline the screening process. Additionally, no crowdsourcing or pre-screened datasets were employed for the records screening.

\subsection{Data collection process}

\noindent For data collection, a single reviewer was responsible for collecting the relevant information from each report. No automation processes were employed for the data collection process.
During the review, all articles were examined in their original language. The selection of articles was based on the predefined eligibility criteria described above. 
No software or automated tools were used to extract data from the figures or graphical representations in the articles. Finally, the data collection process entailed a manual analysis of the articles to extract the pertinent information for the review.

\subsection{Data item}

\noindent In our study, we examined three primary outcomes: segmentation, classification, and registration. All of the reviewed studies were compatible with these three outcome domains. We did not alter or introduce any changes to the outcome domains or their significance in the review. Likewise, we did not modify the selection processes within these eligible outcome domains. Beyond the three outcomes previously mentioned, we also explored data pertaining to the utilization of FSL, OSL, and ZSL techniques, as well as their applications within the field of medical imaging. 

\subsection{Assessment of bias in studies}

\noindent To evaluate the potential risk of bias (ROB) or concerns regarding applicability in each study, we utilized the PROBAST tool \cite{Wolff19}, designed for assessing the quality of diagnostic accuracy studies. For each outcome, we created a table denoting studies with low risk or concerns using a green checkmark symbol \greencheck and those with high risk or concerns using a red cross symbol \redcross. 

\subsection{Effect measures}

\noindent In the segmentation studies included in our review, we evaluated the performance using two popular metrics: the Dice score and the Intersection over Union (IoU). When investigating studies that addressed a classification task, we considered Accuracy, F1-score, Recall, and the Area Under the Receiver Operating Characteristic (AUROC) metrics. Finally, in the registration domain, we examined the Dice score as well as the Average Landmark Distance (ALD) and the Target Registration Error (TRE).

\subsection{Synthesis methods}

\noindent In our systematic review, we structured the key information of each study within dedicated tables for each outcome category. The tables included the following information: first author, year of publication, the algorithm or framework used, the number of training data, the best performance achieved by the model, and whether the study utilized the meta-learning paradigm.  
To provide a visual summary of the results, we used forest plots. We generated these plots by grouping the studies based on the anatomical structure investigated and the meta-learning method employed in each outcome. We created separate forest plots for each performance metric (accuracy, AUROC, etc.), considering, in each study, the highest performance achieved (across various experiments and image modalities). In each forest plot, we reported the mean and the 95\% confidence interval (CI) across all the studies within the corresponding group, whether organized by clinical task or meta-learning algorithm.
It is important to note that we did not conduct a meta-analysis of the collected results. This is because the studies included in our review encompassed various clinical applications, making direct comparisons between the results inappropriate. Therefore, the forest plots served as a visual representation of the individual study findings rather than a quantitative synthesis of the data.
{\color{red}Finally, for each study, we presented a table summarizing the main learning strategies employed w.r.t. three key phases: \textbf{Pre-training}, \textbf{Training}, and \textbf{Data augmentation}. Based on this, we identified a standard methodological pipeline shared among the examined studies.}


\section{Results}

\subsection{Study selection}
\begin{figure*}[h!]
    \centering
    \includegraphics[width=12cm]
    {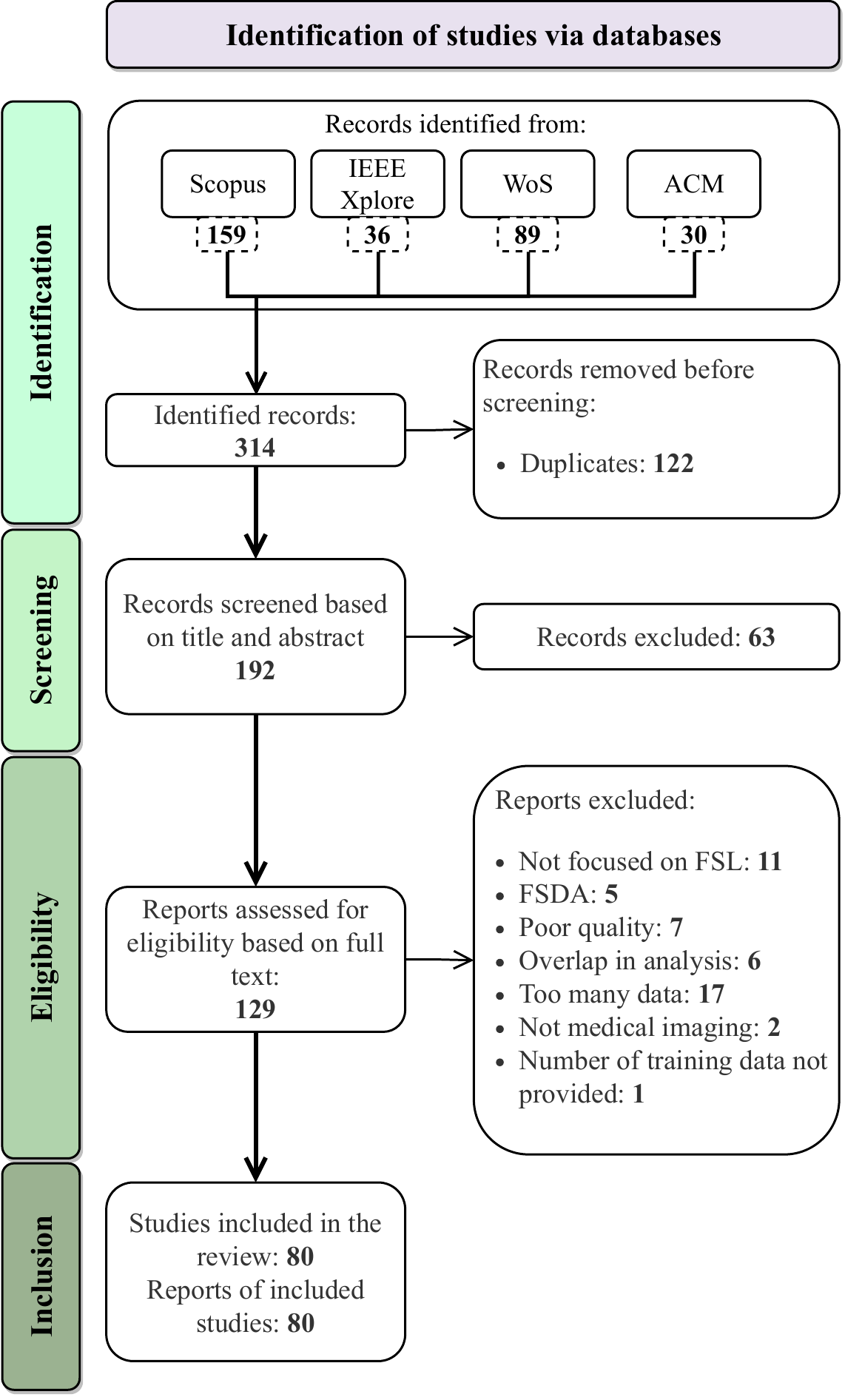}
    \caption{PRISMA flow diagram.}
    \label{fig:prisma_diagram}
\end{figure*}

\noindent In Figure \ref{fig:prisma_diagram}, we summarize the data selection flow using the PRISMA diagram. In total, we retrieved 314 studies and included 80 studies in the final analysis.

\subsection{Studies characteristics}

\begin{figure}[h!]
    \centering
    \includegraphics[width=10cm]{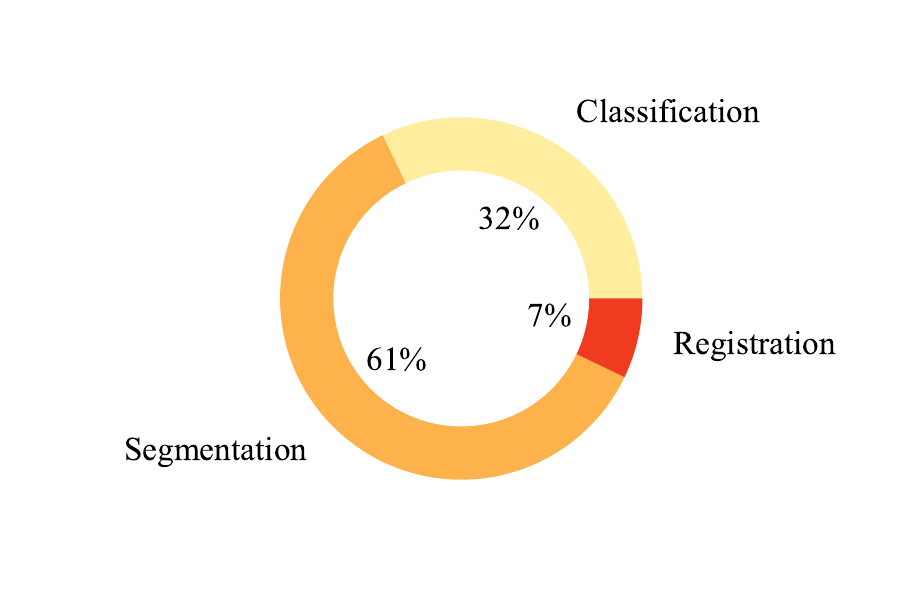}
    \caption{Studies distribution by outcome.}
    \label{fig:outcome_chart}
\end{figure}

\noindent In this section, we present the findings resulting from our analysis of the selected research papers. Figure \ref{fig:outcome_chart} displays the distribution of studies across the three investigated outcomes.
Below, we present the results of our analysis grouped by outcome. It's worth noting that several studies, namely \cite{He20}, \cite{He22}, \cite{Roychowdhury21}, \cite{Shi23}, and \cite{Xu19}, are included multiple times, as they address various outcomes simultaneously.  

\subsubsection{Segmentation}
\noindent We selected 50 relevant studies, each focusing on medical segmentation as its primary task. All information extracted from the selected studies is provided in Table \ref{tab:FSL_segmentation}. In addition, for each study, we present  ROB and the applicability analyses in Table \ref{tab:Bias_FSL_segmentation}. {\color{red} The selected studies implemented different strategies to perform FSL for segmentation purposes. For example, Khadka et al. \cite{Khadka22} leveraged an initialization-based meta-learning approach, specifically the implicit MAML algorithm, to segment skin lesions and colonoscopy images. They evaluated the generalizability of their model by testing it on both the same and a different task w.r.t. the one used during training. Besides meta-learning algorithms, several studies leveraged other types of learning frameworks. For example, Khaled et al. \cite{Khaled22} employed a semi-supervised approach leveraging a multi-stage Generative Adversarial Network (GAN) that performs a coarse-to-fine segmentation of brain tissues. To improve the robustness and generalization capabilities of the model, it is also usual to employ a pre-training step and data augmentation techniques. For example, Xu and Niethammer \cite{Xu19} leveraged a pre-trained network and a joint registration-segmentation network training. Specifically, the registration network provides a realistic form of data augmentation, an approach also used in other FSL segmentation studies.}

\FloatBarrier
\begin{table*}[ht]
\small
\caption{FSL studies for medical image segmentation.\label{tab:FSL_segmentation}}
\centering
\begin{tabular}
{P{.05\textwidth}P{.15\textwidth}P{.25\textwidth}P{.1\textwidth}P{.2\textwidth}P{.15\textwidth}}\\
\toprule
\textbf{Study ID} &\textbf{Pub. ref.} & \textbf{Algorithm/Pipeline} & \textbf{K-shot} & \textbf{Best performance} & \textbf{Meta-learning type} \\
\toprule
1 & \makecell{Blendowski,\\ Nickisch, and \\Heinrich \cite{Blendowski19}} & \makecell{Siamese Network \\ + SSL} & \makecell{1-shot \\ 9-shot} & \makecell{Dice: \\0.853 (Liver)\\ 0.657 (Spleen) \\ 0.663 (Kidney) \\ 0.656 (Psoas) } & None  \\
\midrule
2 & Chan et al. \cite{Chan22} & \makecell{Res2-UNeXT +\\ Data augmentation \\ with Daemons \\ registration algorithm}
&  8-shot	&	\makecell{IoU:\\ 0.943 (Cells)}	& None\\
\midrule
3 & Chen et al. \cite{Chen22} & \makecell{ Adversarial Data \\ Augmentation\\ Framework \\(Advchain)} & \makecell{1-shot \\ 3-shot \\11-shot} & \makecell{Dice: \\ 0.844 (LV) \\ 0.647 (RV) \\ 0.812 (MYO) \\0.572 (Prostate PZ) \\ 0.845 (Prostate CZ) } & None \\
\midrule
4 & Cui et al. \cite{Cui20} &	\makecell{MRE-Net \\(Distance  metric-learning +\\ U-net)}	& \makecell{1-shot \\ 7-shot} & \makecell{Dice: \\ 0.781 (Spleen) \\ 0.774 (Kidney) \\ 0.522 (Gallbladder) \\0.568 (Esophagus) \\ 0.597 (Stomach) \\ 0.613 (Pancreas) \\ 0.820 (Brain, MAS)} & 	Metric learning\\
\midrule
5 & \makecell{Ding, Wangbin \\ et al. \cite{Ding20}} 	&\makecell{Registration model +\\ Similarity model + \\ Patch label fusion}	& 20-shot	& \makecell{Dice: \\0.817 (MYO)} & Metric learning\\
\midrule
6 & Ding, Yu and Yang \cite{Ding21}	&	\makecell{Registration network + \\ Generative network + \\ Segmentation network}&	1-shot	& \makecell{Dice: \\ 0.851 (Brain) }& Hallucination\\
\midrule
7 & Farshad et al. \cite{Farshad22} & \makecell{MetaMedSeg (Reptile-based \\with task weighting)} & 15-shot & \makecell{IoU:\\ 0.683 (Heart) \\ 0.583 (Spleen) \\ 0.227 (Prostate PZ) \\ 0.483 (Prostate TZ)} & Initialization\\
\midrule
8 & Feng et al. \cite{Feng21} & \makecell{Medical Prior-based \\ FSL  Network +\\ Interactive Learning-based \\Test Time Optimization\\ Algorithm }& 10-shot	& \makecell{Dice: \\ 0.569 (Breast)\\ 0.584 (Kidney) \\ 0.751 (Liver) \\ 0.675 (Stomach)}&	Metric learning
\\
\midrule
9 & Gama, Oliveira and dos Santos \cite{Gama21}	&  \makecell{Weakly-supervised \\Segmentation Learning}	& \makecell{1-shot \\5-shot\\10-shot\\20-shot} & \makecell{IoU: \\ 0.870 (Lungs)\\
0.790 (Heart)\\ 0.800 (Mandible) \\ 0.870 (Breast)} &	Initialization
\\
\bottomrule
\end{tabular}
\end{table*}
\FloatBarrier

\FloatBarrier
\begin{table*}[ht]
\small
\ContinuedFloat
\caption{\emph{(continued).} \label{tab:FSL_segmentation}}
\centering
\begin{tabular}
{P{.05\textwidth}P{.15\textwidth}P{.25\textwidth}P{.1\textwidth}P{.2\textwidth}P{.15\textwidth}}\\
\toprule
\textbf{Study ID} &\textbf{Pub. ref.} & \textbf{Algorithm/Pipeline} & \textbf{K-shot} & \textbf{Best performance} & \textbf{Meta-learning type} \\
\toprule
10 & Gama et al. \cite{Gama22} 	& 	ProtoSeg&	\makecell{1-shot \\5-shot\\10-shot\\20-shot} & \makecell{IoU:\\ 0.800 (Heart)\\ 0.920\% (Lungs)\\
0.720\% (Breast) \\ 0.400 (Mandible)} & Metric learning\\
\midrule
11 &  Guo, Odu and Pedrosa \cite{Guo22} & \makecell{Cascaded U-net + \\3D augmentation} & \makecell{From \\ 1-shot\\ to 6-shot} &	\makecell{Dice: \\0.910 (Kidney)} & None\\
\midrule
12 & Hansen et al. \cite{Hansen22} & \makecell{Anomaly detection-inspired\\ model + SSL}&	\makecell{1-shot\\2-shot\\3-shot} & \makecell{Dice: \\ 0.875 (LV-BP) \\ 0.773 (RV) \\ 0.624 (MYO) \\ 0.833 (Kidney) \\ 0.759 (Spleen)\\ 0.808 (Liver)}& Metric learning\\
\midrule
13 & He et al. \cite{He20}&	\makecell{Deep Complementary\\ Joint Model\\ (Segmentation model +\\ Pixel-wise discriminator + \\Registration model)}	& 4-shot &	\makecell{Dice:\\ 0.970 (AA) \\ 0.920 (LA) \\ 0.950 (LV) \\ 0.870 (MYO) \\ 0.800 (PA) \\ 0.800 (RA) \\ 0.810 (RV) }&	None\\
\midrule
14 & He et al. \cite{He22}	& \makecell{Knowledge Consistency\\ Constraint strategy +\\ Space-style Sampling Program + \\Mix Misalignment Regularization} & \makecell{1-shot \\5-shot}	& 	\makecell{Dice: \\  0.911 (Heart, MAS) \\0.872 (Brain, MAS)}&	Hallucination\\
\midrule
15 & Jenssen et al. \cite{Jenssen22} & \makecell{Self-guided \\Anomaly detection-inspired\\ model} & 1-shot & \makecell{Dice:\\ 0.840 (LV) \\ 0.585 (MYO) \\ 0.697 (RV)}  & Metric learning\\
\midrule
16 & Joyce and Kozerke \cite{Joyce19} & \makecell{Anatomical model\\ + SSL}	& \makecell{1-shot \\3-shot\\10-shot}	&	\makecell{Dice: \\ 0.630 (Heart) }&	None\\
\midrule
17 & Khadka et al. \cite{Khadka22} & \makecell{Implicit MAML +\\ Attention U-Net}	& \makecell{5-shot\\10-shot \\20-shot}	&	\makecell{Dice: \\ 0.833 (Skin, nevus)}& Initialization\\
\midrule
18 & Khaled, Han and Ghaleb \cite{Khaled22}& \makecell{Multi-stage GAN}& 	\makecell{5-shot\\10-shot} & \makecell{Dice:\\ 0.940 (Brain, MAS)} &	None\\
\midrule
19 & Khandelwal and Yushkevich \cite{Khandelwal20} &	\makecell{Gradient-based\\ meta-learning\\ domain generalization +\\ 3D U-Net + \\Fine-tuning}& \makecell{2-shot\\4-shot\\6-shot}	& \makecell{Dice: \\ 0.823 (Spine, MAS)} & Initialization\\
\bottomrule
\end{tabular}
\end{table*}
\FloatBarrier


\FloatBarrier
\begin{table*}[ht]
\ContinuedFloat
\small
\caption{\emph{(continued)}.\label{tab:FSL_segmentation}}
\centering
\begin{tabular}
{P{.05\textwidth}P{.15\textwidth}P{.25\textwidth}P{.1\textwidth}P{.2\textwidth}P{.15\textwidth}}\\
\toprule
\textbf{Study ID} &\textbf{Pub. ref.} & \textbf{Algorithm/Pipeline} & \textbf{K-shot} & \textbf{Best performance}  & \textbf{Meta-learning \ type} \\
\toprule
20 & Kim et al. \cite{Kim21} & \makecell{VGG16 + \\ Bidirectional gated \\recurrent unit +\\ U-Net + Fine-tuning} & 5-shot	& \makecell{Dice:\\ 0.905 (Spleen)\\ 0.900 (Kidney) \\ 0.887 (Liver) \\ 0.771 (Bladder)}& Metric learning
\\
\midrule
21 & Li et al. \cite{Li22}	&\makecell{3D U-Net + \\Prototypical learning +\\ Image alignment module} &	1-shot	&	\makecell{Dice: \\ 0.417 (Prostate, MAS)}&	Metric learning\\
\midrule
22 & Lu et al. \cite{Lu20} &	\makecell{Contour Transformer Network \\(ResNet-50 + \\Graph convolutional\\ network blocks)} &	1-shot &	\makecell{IoU:\\ 0.973 (Knee)\\ 0.948 (Lung)\\ 0.970 (Phalanx)\\ 0.973 (Hip)}&	None\\
\midrule
23 & Lu and Ye \cite{Lu21}	& \makecell{TractSeg +\\ Knowledge transfer\\ with warmup} &	\makecell{1-shot\\ 5-shot}	& \makecell{Dice: \\ 0.812 \\(Brain, WM) }& None\\
\midrule
24 & Ma et al. \cite{Ma21} &\makecell{Segmentation network +\\ Zero-shot segmentation\\ network + \\Spatial Context\\ Attention module}& 0-shot	& \makecell{Dice: \\ 0.882 \\(Brain, tumour)}& None \\
\midrule
25 & Niu et al. \cite{Niu22} & \makecell{Conditioner + Segmenter + \\Symmetrical Supervision\\ Mechanism +\\ Transformer-based Global \\Feature Alignment module} & 1-shot & \makecell{Dice: \\  0.870 (LV-BP) \\ 0.815 (Kidney) \\ 0.738 (Spleen) \\0.729 (Liver)} & Metric learning
 \\
\midrule
26 & Ouyang et al. \cite{Ouyang22}  & \makecell{Self-Supervised Adaptive\\ Local Prototype \\Pooling Network} &\makecell{1-shot \\ 5-shot} &	\makecell{Dice:\\ 0.862 (Kidney)\\ 0.757 (Spleen) \\ 0.821 (Liver) \\ 0.870 (LV)\\ 0.721 (MYO) \\ 0.860 (RV) } &	Metric learning\\
\midrule
27 & Pham et al. \cite{Pham20} & \makecell{Few-Sample-Fitting} & \makecell{1-shot to \\ 20-shot}	&	\makecell{Dice: \\ 0.990 (Femur)}&None\\
\midrule
28 &  Pham, Dovletov and Pauli \cite{Pham21} &	\makecell{3D U-Net + \\Imitating encoder +\\ Prior encoder + \\Joint decoder}	& 1-shot	& \makecell{Dice:\\ 0.776 (Liver)}&	None\\
\midrule
29 & Roy et al. \cite{Roy20}	& \makecell{Conditioner arm + \\Segmenter arm +\\ Channel Squeeze \& \\ Spatial Excitation\\ blocks} & 1-shot	&\makecell{Dice:\\ 0.700 (Liver) \\0.607 (Spleen)\\ 0.464 (Kidney)\\ 0.499 (Psoas)}&	None\\
\midrule
30 & Roychowdhury et al. \cite{Roychowdhury21} &	\makecell{Echo state network +\\ augmented U-Net}	& 5-shot & \makecell{Dice: \\ 0.640 (Eye, IC)} &	None\\
\bottomrule
\end{tabular}
\end{table*}
\FloatBarrier


\FloatBarrier
\begin{table*}[ht]
\ContinuedFloat
\small
\caption{\emph{(continued).}\label{tab:FSL_segmentation}}
\centering
\begin{tabular}
{P{.05\textwidth}P{.15\textwidth}P{.25\textwidth}P{.1\textwidth}P{.2\textwidth}P{.15\textwidth}}\\
\toprule
\textbf{Study ID} &\textbf{Pub. ref.} & \textbf{Algorithm/Pipeline} & \textbf{K-shot} & \textbf{Best performance} & \textbf{Meta-learning type} \\
\toprule
31 & Rutter, Lagergren and Flores \cite{Rutter19} &		\makecell{CNN for\\ Boundary Optimization}& \makecell{1-shot \\3-shot\\5-shot} &\makecell{Dice: \\ 0.931 (Cells)}&	None\\
\midrule
32 & Shen et al. \cite{Shen20}	& \makecell{Large Deformation\\ Diffeomorphic Metric \\Mapping model + \\Sample transformations + \\Interpolation}	& 1-shot	& \makecell{Dice:\\ 0.883 (Knee)}	& None\\
\midrule
33 & Shen et al. \cite{Shen21} &  \makecell{VGG-16 + \\Poisson learning + \\Spatial Consistency \\Calibration} & 1-shot &	\makecell{Dice: \\ 0.619 (Skin, MAD) \\ 0.610 (Liver) \\ 0.536 (Kidney) \\ 0.529 (Spleen)}& None\\
\midrule
34 & Shi et al. \cite{Shi23} & \makecell{Joint Registration \\and Segmentation\\ Self-training\\ Framework (JRSS)} & 5-shot & \makecell{Dice: \\0.795 \\(Brain, MAS) \\ 0.753\\ (Abdomen, MAS)} & None \\
\midrule
35 & Sun et al. \cite{Sun22}&  \makecell{2-branch CNN  + \\Spatial Squeeze \\ Excite module + \\Global Correlation module +\\ Discriminative Embedding\\ module}	& \makecell{1-shot} & \makecell{Dice: \\ 0.495 (Liver) \\ 0.606 (Spleen) \\ 0.830 (Kidney)} & Metric learning\\
\midrule
36 & Tang et al. \cite{Tang21}	& \makecell{Recurrent Prototypical\\ Networks (U-Net + \\Contex Relation Encoder + \\Prototypical Network)}&	\makecell{1-shot}	& \makecell{Dice: \\ 0.788 (Spleen)\\ 0.851 (Kidney)\\ 0.819 (Liver)}& Metric learning\\
\midrule
37 & Tomar et al. \cite{Tomar22}	& 	\makecell{Generative Style Transfer \\(Appearance model + \\Style encoder +\\ Flow model + \\ Flow Adversarial\\ Autoencoder)} &	1-shot &	\makecell{Dice:\\ 0.835 (Brain, MAS)}& None\\
\midrule
38 & Wang et al. \cite{Wang20} &	\makecell{Label Transfer \\ Network \\(Atlas-based \\segmentation + \\ Forward-backward\\ correspondance)}	& 1-shot	&	\makecell{Dice: \\0.823 (Brain, MAS)}& None\\
\midrule
39 & Wang et al. \cite{Wang21} & \makecell{Siamese model and \\ Individual-Difference-\\Aware model (Encoders +\\ Forward-backward\\ consistency)}& \makecell{1-shot \\5-shot}&	\makecell{Dice: \\ 0.862 (Brain, MAS)\\ 0.803 (Spleen) \\ 0.884 (Kidney) \\ 0.916 (Liver) \\ 0.684 (Stomach) \\ 0.511 (Pancreas) \\ 0.485 (Doudenum) \\ 0.519 (Esophagus) }	& None
\\
\bottomrule
\end{tabular}
\end{table*}
\FloatBarrier

\FloatBarrier
\begin{table*}[ht]
\ContinuedFloat
\small
\caption{\emph{(continued).}\label{tab:FSL_segmentation}}
\centering
\begin{tabular}
{P{.05\textwidth}P{.15\textwidth}P{.25\textwidth}P{.1\textwidth}P{.2\textwidth}P{.15\textwidth}}\\
\toprule
\textbf{Study ID} &\textbf{Pub. ref.} & \textbf{Algorithm/Pipeline} & \textbf{K-shot} & \textbf{Best performance} & \textbf{Meta-learning type}\\
\toprule
40 & Wang et al. \cite{Wang221}& \makecell{V-Net + \\Init-crop +\\ Self-down + \\Self-crop} &	4-shot &	\makecell{Dice: \\ 0.937 (LV) \\0.890 (RV) \\ 0.872 (LA) \\ 0.909 (RA) \\ 0.831 (MYO) \\0.943 (AO) \\0.798 (PA) }	& None\\
\midrule
41 & Wang, Zhou and Zheng \cite{Wang322}  & \makecell{Prototype learning + \\Self-reference + \\Contrastive learning} & 1-shot & \makecell{Dice: \\0.756 (Liver)\\ 0.737 (Spleen)\\ 0.842 (Kidney)} & Metric learning\\
\midrule
42 &Wang et al. \cite{Wang22} & 	\makecell{Alternating Union Network\\ (Image Sub-Network + \\ Label Sub-Network}	& 1-shot	&	\makecell{Dice:\\  0.873 (LV) \\ 0.637 (MYO) \\ 0.720 (RV) } &	None\\
\midrule
43 & Wu, Xiao and Liang \cite{Wu22} & \makecell{Dual Contrastive Learning + \\Anatomical Auxiliary\\ Supervision + \\Constrained Iterative \\Prediction module}& 1-shot & \makecell{Dice: \\0.699 (Liver)\\ 0.838 (Kidney)\\ 0.749 (Spleen)} & None \\
\midrule
44 & Wu et al. \cite{Wu222} & \makecell{Self-Learning + \\ One-Shot\\ Learning} & 1-shot & \makecell{Dice: \\ 0.850 (Spleen) \\ 0.930 (Liver)} & None \\ 
\midrule
45 & Xu and Niethammer \cite{Xu19} &	\makecell{DeepAtlas \\(Semi-Supervised\\ Learning + \\Segmentation network + \\Registration network)}	& \makecell{1-shot \\ 5-shot\\10-shot} &	\makecell{Dice: \\ 
 0.892 (Knee, MAS)\\0.612 (Brain)\\ } &	None
\\
\midrule
46 & Yu et al. \cite{Yu21} 	& \makecell{Location-Sensitive\\ Local Prototype \\Network}	& 1-shot	& \makecell{Dice: \\ 0.793 (Liver)\\ 0.733 (Spleen) \\  0.765 (Kidney) \\0.524 (Psoas)}	& Metric learning\\
\midrule
47 & Yuan, Esteva and Xu \cite{Yuan21}  &\makecell{	MetaHistoSeg\\ (U-Net + MAML)} &	8-shot	& \makecell{IoU: \\ 0.326 (Cells) \\ 0.682 (Cells nuclei) \\ 0.557 (Gland) \\ 0.632 (Colon, tumour)} &	Initialization
\\
\midrule
48 & Zhao et al. \cite{Zhao19} & \makecell{Spatial and appearance \\transform models + \\Semi-supervised learning + \\ Supervised learning}	& 1-shot	& \makecell{Dice: \\0.815 (Brain, MAS)}	& None\\
\midrule
49 & Zhao et al. \cite{Zhao22} & Meta-hallucinator & \makecell{1-shot \\ 4-shot} & \makecell{Dice:\\ 0.756 (AO) \\0.751 (LA) \\ 0.823 (LV) \\ 0.696 (MYO) }&Initialization and Hallucination-based \\
\bottomrule
\end{tabular}
\end{table*}
\FloatBarrier


\FloatBarrier
\begin{table*}[ht]
\ContinuedFloat
\small
\caption{\emph{(continued).}\label{tab:FSL_segmentation}}
\centering
\begin{tabular}
{P{.05\textwidth}P{.15\textwidth}P{.25\textwidth}P{.1\textwidth}P{.2\textwidth}P{.15\textwidth}}\\
\toprule
\textbf{Study ID} &\textbf{Pub. ref.} & \textbf{Algorithm/Pipeline} & \textbf{K-shot} & \textbf{Best performance} & \textbf{Meta-learning type}\\
\toprule
50 & Zhou et al. \cite{Zhou21} &	\makecell{OrganNet \\(3 encoders +\\ Pyramid Reasoning\\ Modules)}& 1-shot	& \makecell{Dice: \\0.891 (Spleen)\\ 0.860 (Kidney) \\0.770 (Aorta)\\ 0.728 (Pancreas) \\0.826 (Stomach)}&	None\\
\bottomrule
\end{tabular}
\end{table*}
\normalsize
\FloatBarrier

\FloatBarrier
\begin{table*}[ht]
\small
\caption{ROB of FSL studies for medical image segmentation.}\label{tab:Bias_FSL_segmentation}
\centering
\begin{tabular}
{P{.05\textwidth}P{.2\textwidth}P{.05\textwidth}P{.05\textwidth}P{.05\textwidth}P{.08\textwidth}P{.08\textwidth}P{.05\textwidth}P{.05\textwidth}P{.05\textwidth}P{.05\textwidth}}\\
\toprule
&& \multicolumn{5}{c}{\textbf{Risk of Bias}} & \multicolumn{4}{c}{\textbf{Applicability}}\\
\cmidrule(lr){3-7}
\cmidrule(lr){8-11}
\textbf{Study ID}&	\textbf{Pub. ref.} &\textbf{Part.}& \textbf{Pred} & \textbf{Out.} & \textbf{Analysis} & \textbf{Overall} & \textbf{Part.} & \textbf{Pred.} & \textbf{Out.} & \textbf{Overall}\\
\toprule
1	& Blendowski, Nickisch, and Heinrich \cite{Blendowski19}	& \greencheck	& \greencheck	& \greencheck	& \greencheck	& \greencheck	& \greencheck	& \greencheck	& \greencheck	& \greencheck\\
\midrule
2	& Chan et al. \cite{Chan22}	&\greencheck	& \greencheck	& \greencheck	& \redcross	& \redcross	& \greencheck	& \greencheck	& \greencheck	& \greencheck\\
\midrule
3	& Chen et al. \cite{Chen22}	&\greencheck	& \greencheck	& \greencheck	& \greencheck	& \greencheck	& \greencheck	& \greencheck	& \greencheck	& \greencheck\\
\midrule
4	& Cui et al. \cite{Cui20}&	\greencheck	& \greencheck	& \greencheck	& \greencheck	& \greencheck	& \greencheck	& \greencheck	& \greencheck	& \greencheck\\
\midrule
5	& Ding, Wangbin et al. \cite{Ding20}	&\greencheck	& \greencheck	& \greencheck	& \redcross	& \redcross	& \greencheck	& \greencheck & \greencheck	&\greencheck\\
\midrule
6	& Ding, Yu and Yang \cite{Ding21} 	&\greencheck	& \greencheck	& \greencheck	& \greencheck	& \greencheck	& \greencheck	&\greencheck &	\greencheck	& \greencheck\\
\midrule
7	&Farshad et al. \cite{Farshad22}&	\greencheck	& \greencheck	& \greencheck	& \redcross	& \redcross	& \greencheck	& \greencheck	& \greencheck	& \greencheck\\
\midrule
8	& Feng et al. \cite{Feng21}	& \greencheck	& \greencheck	& \greencheck	& \greencheck	& \greencheck	& \greencheck	& \greencheck	& \greencheck	& \greencheck\\
\midrule
9	& Gama, Oliveira and dos Santos \cite{Gama21} &	\greencheck	& \greencheck	& \greencheck	& \greencheck	& \greencheck	& \greencheck	& \greencheck	& \greencheck	& \greencheck\\
\midrule
10	& Gama et al. \cite{Gama22} &	\greencheck	& \greencheck	& \greencheck	& \greencheck	& \greencheck	& \greencheck	& \greencheck	& \greencheck	& \greencheck\\
\midrule
11	&  Guo, Odu and Pedrosa \cite{Guo22} &	\greencheck	& \greencheck	& \greencheck	& \redcross	& \redcross	& \greencheck	& \greencheck	& \greencheck	& \greencheck\\
\midrule
12	& Hansen et al. \cite{Hansen22}	&\greencheck	& \greencheck	& \greencheck	& \greencheck	& \greencheck	& \greencheck	& \greencheck	& \greencheck	& \greencheck\\
\midrule
13& 	He et al. \cite{He20} &	\greencheck	& \greencheck	& \greencheck	& \redcross	& \redcross	& \greencheck&	\greencheck&	\greencheck&	\greencheck\\
\midrule
14	& He et al. \cite{He22} &	\greencheck	& \greencheck	& \greencheck	& \greencheck	& \greencheck	& \greencheck	& \greencheck	& \greencheck	& \greencheck\\
\midrule
15	& Jenssen et al. \cite{Jenssen22}	& \greencheck	& \greencheck	& \greencheck	& \redcross	& \redcross	& \greencheck	& \greencheck	& \greencheck	& \greencheck\\
\midrule
16	& Joyce and Kozerke \cite{Joyce19} & 	\greencheck	& \greencheck	& \greencheck	& \redcross	& \redcross	& \greencheck	& \greencheck	& \greencheck	& \greencheck\\
\midrule
17&	Khadka et al. \cite{Khadka22}	&\greencheck	& \greencheck	& \greencheck	& \greencheck	& \greencheck	& \greencheck	& \greencheck	& \greencheck	& \greencheck\\
\midrule
18&	Khaled, Han and Ghaleb \cite{Khaled22}	&\greencheck	& \greencheck	& \greencheck	& \greencheck	& \greencheck	& \greencheck	& \greencheck	& \greencheck	& \greencheck\\
\bottomrule
\end{tabular}
\end{table*}

\begin{table*}[ht]
\ContinuedFloat
\small
\caption{\emph{(continued)}.}\label{tab:Bias_FSL_segmentation}
\centering
\begin{tabular}[!t]
{P{.05\textwidth}P{.2\textwidth}P{.05\textwidth}P{.05\textwidth}P{.05\textwidth}P{.08\textwidth}P{.08\textwidth}P{.05\textwidth}P{.05\textwidth}P{.05\textwidth}P{.05\textwidth}}\\
\toprule
&& \multicolumn{5}{c}{\textbf{Risk of Bias}} & \multicolumn{4}{c}{\textbf{Applicability}}\\
\cmidrule(lr){3-7}
\cmidrule(lr){8-11}
\textbf{Study ID}&	\textbf{Pub. ref.} &\textbf{Part.}& \textbf{Pred} & \textbf{Out.} & \textbf{Analysis} & \textbf{Overall} & \textbf{Part.} & \textbf{Pred.} & \textbf{Out.} & \textbf{Overall}\\
\toprule
19	& Khandelwal and Yushkevich \cite{Khandelwal20}	& \greencheck	& \greencheck	& \greencheck	& \redcross	& \redcross	& \greencheck	   &\greencheck	& \greencheck&	\greencheck\\
\midrule
20&	Kim et al. \cite{Kim21}	& \greencheck	& \greencheck	& \greencheck	& \greencheck	& \greencheck	& \greencheck	& \greencheck	& \greencheck	& \greencheck\\
\midrule
21	& Li et al. \cite{Li22} &	\greencheck	& \greencheck	& \greencheck	& \greencheck	& \greencheck	& \greencheck	& \greencheck	& \greencheck	& \greencheck\\
\midrule
22	& Lu et al. \cite{Lu20}	& \greencheck	& \greencheck	& \greencheck	& \greencheck	& \greencheck	& \greencheck	& \greencheck	& \greencheck	& \greencheck\\
\midrule
23 & Lu and Ye \cite{Lu21} &	\greencheck	& \greencheck	& \greencheck	& \greencheck	& \greencheck	& \greencheck	& \greencheck	& \greencheck	& \greencheck\\
\midrule
24	& Ma et al. \cite{Ma21} &	\greencheck	& \greencheck	& \greencheck	& \greencheck	& \greencheck	& \greencheck	& \greencheck	& \greencheck	& \greencheck\\
\midrule
25 &	Niu et al. \cite{Niu22}	& \greencheck	& \greencheck	& \greencheck	& \greencheck	& \greencheck	& \greencheck	& \greencheck	& \greencheck	& \greencheck\\
\midrule
26	& Ouyang et al. \cite{Ouyang22}	& \greencheck	& \greencheck	& \greencheck	& \greencheck	& \greencheck	& \greencheck	& \greencheck	& \greencheck	& \greencheck\\
\midrule
27	& Pham et al. \cite{Pham20}&\greencheck	& \greencheck	& \greencheck	& \redcross	& \redcross	& \greencheck	& \greencheck	& \greencheck	& \greencheck\\
\midrule
28	& Pham, Dovletov and Pauli \cite{Pham21}	& \greencheck	& \greencheck	& \greencheck	& \greencheck	& \greencheck	& \greencheck	& \greencheck	& \greencheck	& \greencheck\\
\midrule
29	& Roy et al. \cite{Roy20} &\greencheck	& \greencheck	& \greencheck	& \greencheck	& \greencheck	& \greencheck	& \greencheck	& \greencheck	& \greencheck\\
\midrule
30	& Roychowdhury et al. \cite{Roychowdhury21}	& \greencheck	& \greencheck	& \greencheck	& \redcross	& \redcross	& \greencheck	& \greencheck &	\greencheck	& \greencheck\\
\midrule
31	& Rutter, Lagergren and Flores \cite{Rutter19}	& \greencheck	& \greencheck	& \greencheck	& \greencheck	& \greencheck	& \greencheck	& \greencheck	& \greencheck	& \greencheck\\
\midrule
32	& Shen et al. \cite{Shen20}	& \greencheck	& \greencheck	& \greencheck	& \greencheck	& \greencheck	& \greencheck	& \greencheck	& \greencheck	& \greencheck\\
\midrule
33	& Shen et al. \cite{Shen21}	& \greencheck	& \greencheck	& \greencheck	& \greencheck	& \greencheck	& \greencheck	& \greencheck	& \greencheck	& \greencheck\\
\midrule
34	& Shi et al. \cite{Shi23}	& \greencheck	& \greencheck	& \greencheck	& \greencheck	& \greencheck	& \greencheck	& \greencheck	& \greencheck	& \greencheck\\
\midrule
35 & 	Sun et al. \cite{Sun22}	& \greencheck	& \greencheck	& \greencheck	& \greencheck	& \greencheck	& \greencheck	& \greencheck	& \greencheck	& \greencheck\\
\midrule
36	& Tang et al. \cite{Tang21}	& \greencheck	& \greencheck	& \greencheck	& \greencheck	& \greencheck	& \greencheck	& \greencheck	& \greencheck	& \greencheck\\
\midrule
37	& Tomar et al. \cite{Tomar22}	& \greencheck	& \greencheck	& \greencheck	& \greencheck	& \greencheck	& \greencheck	& \greencheck	& \greencheck	& \greencheck\\
\midrule
38	& Wang et al. \cite{Wang20}	& \greencheck	& \greencheck	& \greencheck	& \greencheck	& \greencheck	& \greencheck	& \greencheck	& \greencheck	& \greencheck\\
\midrule
39	& Wang et al. \cite{Wang21}	& \greencheck	& \greencheck	& \greencheck	& \greencheck	& \greencheck	& \greencheck	& \greencheck	& \greencheck	& \greencheck\\
\midrule
40 &	Wang et al. \cite{Wang221}	& \greencheck	& \greencheck	& \greencheck	& \greencheck	& \greencheck	& \greencheck	& \greencheck	& \greencheck	& \greencheck\\
\midrule
41	& Wang, Zhou and Zheng \cite{Wang322}	& \greencheck	& \greencheck	& \greencheck	& \greencheck	& \greencheck	& \greencheck	& \greencheck	& \greencheck	& \greencheck\\
\midrule
42	& Wang et al. \cite{Wang22}	& \greencheck	& \greencheck	& \greencheck	& \greencheck	& \greencheck	& \greencheck	& \greencheck	& \greencheck	& \greencheck\\
\midrule
43	& Wu, Xiao and Liang \cite{Wu22}	& \greencheck	& \greencheck	& \greencheck	& \greencheck	& \greencheck	& \greencheck	& \greencheck	& \greencheck	& \greencheck\\
\midrule
44	& Wu et al. \cite{Wu222}	& \greencheck	& \greencheck	& \greencheck	& \greencheck	& \greencheck	& \greencheck	& \greencheck	& \greencheck	& \greencheck\\
\midrule
45	& Xu and Niethammer \cite{Xu19}	& \greencheck	& \greencheck	& \greencheck	& \greencheck	& \greencheck	& \greencheck	& \greencheck	& \greencheck	& \greencheck\\
\midrule
46	& Yu et al. \cite{Yu21}	& \greencheck	& \greencheck	& \greencheck	& \greencheck	& \greencheck	& \greencheck	& \greencheck	& \greencheck	& \greencheck\\
\midrule
47	& Yuan, Esteva and Xu \cite{Yuan21}.&	\greencheck	& \greencheck	& \greencheck	& \greencheck	& \greencheck	& \greencheck	& \greencheck	& \greencheck	& \greencheck\\
\midrule
48	& Zhao et al. \cite{Zhao19}	& \greencheck	& \greencheck	& \greencheck	& \greencheck	& \greencheck	& \greencheck	& \greencheck	& \greencheck	& \greencheck\\
\midrule
49	& Zhao et al. \cite{Zhao22}	& \greencheck	& \greencheck	& \greencheck	& \greencheck	& \greencheck	& \greencheck	& \greencheck	& \greencheck	& \greencheck\\
\midrule
50	& Zhou et al. \cite{Zhou21}.	& \greencheck	& \greencheck	& \greencheck	& \greencheck	& \greencheck	& \greencheck	& \greencheck	& \greencheck	& \greencheck\\
\bottomrule
\end{tabular}
\end{table*}
\FloatBarrier

\noindent {\color{red} Here, we present the findings of our segmentation papers analysis.}

\begin{figure*}[ht]
\centering
\subfloat[Grouped by clinical task]{\includegraphics[width=0.5\textwidth]{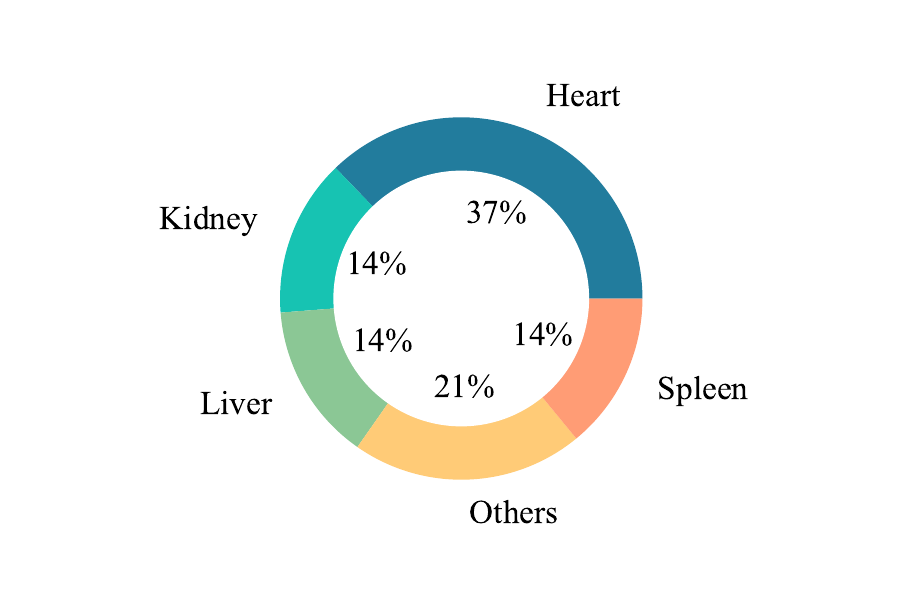}
\label{fig:chart_seg_organ}}
\subfloat[Grouped by meta-learning method]{\includegraphics[width=0.5\textwidth]{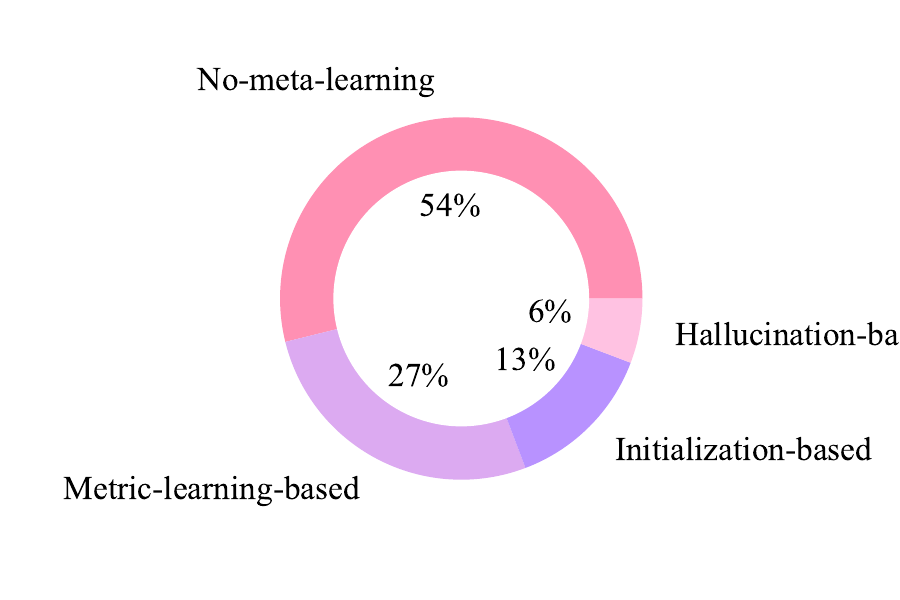}
\label{fig:chart_seg_method}}
\caption{Segmentation studies distribution grouped by clinical task and meta-learning method.}
\label{fig:charts_seg}
\end{figure*}

\noindent\textbf{\color{red}{Clinical task.}}
The segmentation studies investigated various anatomical structures and regions, as well as specific diseases such as polyps or tumours. Here's a breakdown of the papers categorized w.r.t. the clinical task addressed. Eighteen papers (36\%) focused on liver segmentation; 18 studies (36\%) concentrated on kidney segmentation; 17 papers (34\%) centred around spleen segmentation; three papers (6\%) pertained to psoas segmentation; four (8\%) investigated a prostate segmentation task; three works (6\%) involved bladder segmentation; four papers (8\%) dealt with breast segmentation; one paper (2\%) addressed colon segmentation; six (12\%) concerned stomach segmentation; 12 (24\%) were dedicated to brain segmentation; 14 papers (28\%) revolved around heart segmentation; three (6\%) involved pancreas segmentation; three (6\%) pertained to cell segmentation; two papers (4\%) considered to lung segmentation; one (2\%) focused on eye segmentation; two papers (4\%) involved mandible segmentation; one (2\%) addressed duodenum segmentation; two papers (4\%) dealt with skin segmentation; three papers (6\%) considered to knee segmentation; one (2\%) concerned phalanx segmentation; one (2\%) dealt with hip segmentation; one paper (2\%) was dedicated to spine segmentation.
Figure \ref{fig:chart_seg_organ} illustrates the distribution of segmentation studies among the five most popular clinical tasks.

\noindent\textbf{\color{red}{Meta-learning method.}} Out of the 50 studies we selected in the realm of FSL for medical image segmentation, the distribution of their meta-learning methods is as follows: six studies (12\%) leveraged initialization-based methods; 14 studies (28\%) utilized metric learning-based techniques; three studies (6\%) employed hallucination-based methods; one study (2\%) combined both initialization-based and hallucination-based methods. The remaining 28 studies (56\%) did not incorporate any meta-learning technique.
For a visual representation of the distribution of studies in terms of the meta-learning approach employed, refer to Figure \ref{fig:chart_seg_method}.

\noindent\textbf{K-shot.}
According to the k-shot configuration used, we found that 15 studies (30\%) utilized k-shot training with k ranging from 2 to 20; 14 studies (28\%) performed both OSL and FSL; 20 works (40\%) exclusively focused on OSL and one paper (2\%) employed ZSL.

\noindent\textbf{\color{red}{Image modality.}}
{\color{red} In the following, we provide the distribution of studies in terms of the imaging modalities utilized:} 26 (52\%) used CT images; 30 papers (60\%) utilized MRI; four (8\%) relied on X-ray images; two (4\%) involved dermoscopic images; one paper (2\%) made use of endoscopic images; one (2\%) used histopathology images; two (4\%) employed microscopic images; one paper (2\%) utilized OCT images.

\noindent\textbf{Model evaluation.}
For evaluating the models' robustness, the selected studies used different evaluation techniques: 21 studies (42\%) exclusively conducted ablation studies; 11 studies (22\%) utilized both ablation studies and cross-validation; five studies (10\%) relied solely on cross-validation; 13 studies (26\%) did not employ any specific model evaluation technique.

\begin{figure*}[!h]
    \centering
    \includegraphics[width=13cm]{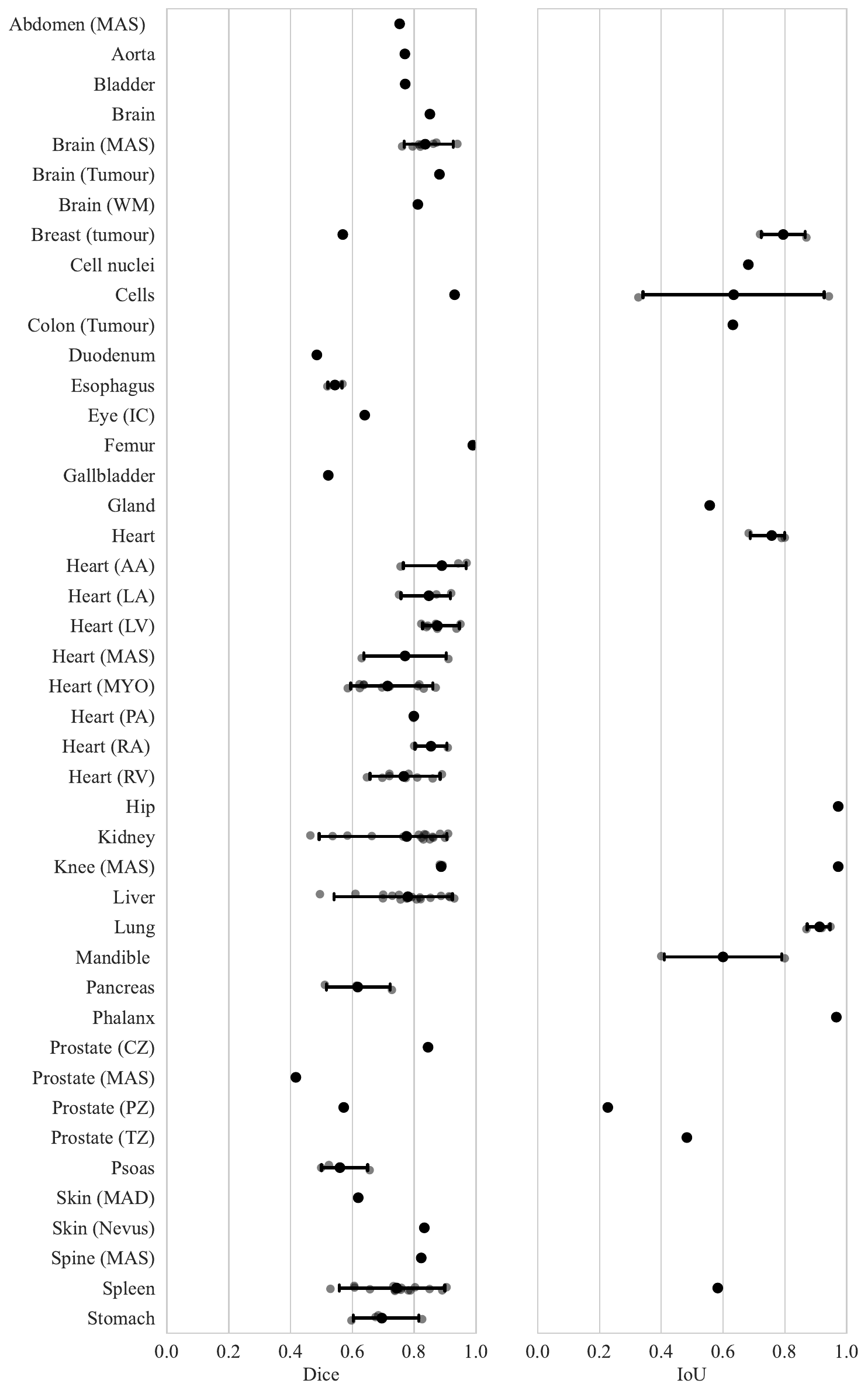}
    \caption{Forest plot of segmentation studies performance based on Dice and IoU metrics. Studies are grouped by the clinical task investigated.  AA = Ascending Aorta; IC = Intraretinal Cyst; LA = Left Atrium; LV = Left Ventricle; MAS = Mean Across Structures; MYO = Myocardium; PA = Pulmonary Artery; PZ = Peripheral Zone; RA = Right Atrium; RV = Right Ventricle; TZ = Transitional Zone; WM = White Matter.}
    \label{fig:fp_seg_organ}
\end{figure*}

\begin{figure*}[!h]
    \centering
    \includegraphics[width=15cm]{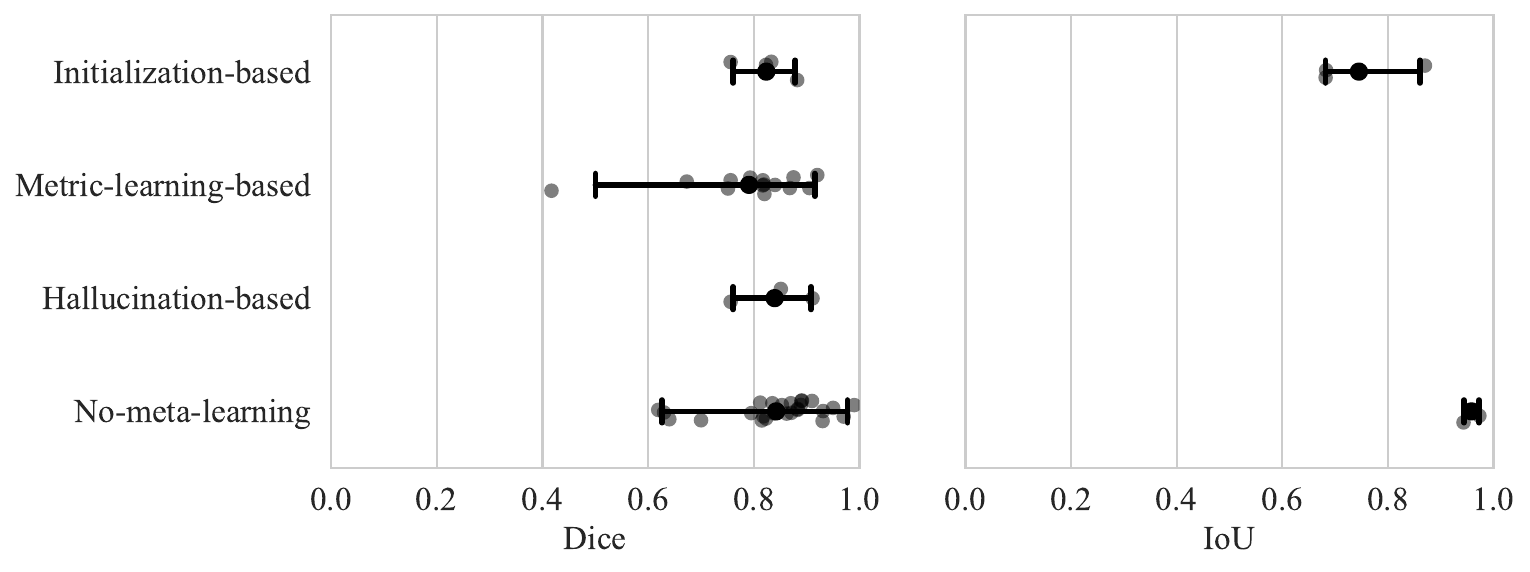}
    \caption{Forest plot of segmentation studies performance based on Dice and IoU metrics. Studies are grouped by the meta-learning method employed.}
    \label{fig:fp_seg_method}
\end{figure*}

\noindent\textbf{\color{red}{Statistical analysis.}} {\color{red}Figure~\ref{fig:fp_seg_organ} displays forest plots that summarize the performance of FSL models (mean and 95\% CI) for each clinical task considering both Dice score and IoU metrics.
Conversely, Figure~\ref{fig:fp_seg_method} depicts the models' results grouped by the meta-learning method used.}

\noindent\textbf{{\color{red} Standard pipeline.}} {\color{red} Table~\ref{tab:main_pipeline_table_seg} details the learning frameworks employed in each step of the standard pipeline by the segmentation studies analyzed in this review. In the following, we explore the distribution of the studies based on the techniques employed for each pipeline phase:} two out of 50 studies (4\%) employed meta-learning for pre-training; two studies (4\%) utilized self-supervised learning and 13 studies (26\%) relied on supervised learning. The majority, 33 out of 50 studies (66\%), did not employ any pre-training stage. For their main training stage, 20 studies (40\%) utilized meta-learning methods; 12 (24\%) employed semi-supervised approaches; four studies (8\%) employed self-supervised methods; 16 studies (32\%) used traditional supervised techniques; one study (2\%) employed a zero-shot learning method. Finally, concerning the data augmentation techniques, 16 studies (32\%) exploited classical data augmentation techniques; five studies (10\%) utilized generative methods for data augmentation; nine studies (18\%) relied on registration-based augmentation. The remaining 24 out of 50 studies (48\%) did not employ data augmentation.

\FloatBarrier
\begin{table*}[ht]
\caption{Standard pipeline steps adopted by segmentation studies.\label{tab:main_pipeline_table_seg}}
\small
\centering
\begin{tabular}
{P{.1\textwidth}P{.2\textwidth}P{.25\textwidth}P{.15\textwidth}P{.2\textwidth}}\\
\toprule
\textbf{Study ID} & \textbf{Pub. ref.} & \textbf{Pre-training} & \textbf{Training}&\textbf{Data augmentation}\\
\toprule
1 & Blendowski, Nickisch, and Heinrich \cite{Blendowski19} & Self-supervised & None & None\\
2 & Chan et al. \cite{Chan22} & None & Supervised & Registration-based\\
3 & Chen et al. \cite{Chen22} &  None & Supervised & Generative\\
4 & Cui et al. \cite{Cui20} & None & Meta & Classical\\
5 & Ding, Wangbin et al.	\cite{Ding20} & None & Meta & Classical \\
6 & Ding, Yu and Yang \cite{Ding21} & None & Semi-supervised and Meta & Generative and Registration-based\\
7 & Farshad et al. \cite{Farshad22} & Meta &  Supervised & None\\
8 & Feng et al. \cite{Feng21} & None & Meta & None \\
9 & Gama, Oliveira and dos Santos \cite{Gama21} & None & Meta & None \\
10 & Gama et al. \cite{Gama22} & None & Meta & None \\
11 & Guo, Odu and Pedrosa \cite{Guo22} & None & Supervised & Classical \\
12 & Hansen et al. \cite{Hansen22} & Supervised & Self-supervised and Meta & None \\
13 & He et al. \cite{He20} & None & Supervised & Registration-based \\
14 & He et al. \cite{He22} & None & Meta & Registration-based\\
15 & Jenssen et al. \cite{Jenssen22} & Supervised & Self-supervised and Meta & None \\
16 & Joyce and Kozerke \cite{Joyce19} & None & Self-supervised & Classical \\
17 & Khadka et al. \cite{Khadka22} & Supervised & Meta & None \\
18 & Khaled, Han and Ghaleb \cite{Khaled22} & None & Semi-supervised & None \\
19 & Khandelwal and Yushkevich \cite{Khandelwal20} & None & Meta & Classical \\
20 & Kim et al. \cite{Kim21} & Meta & Meta & Classical \\
21 & Li et al. \cite{Li22} & None & Meta & Classical \\
22& Lu et al. \cite{Lu20} & Supervised & Semi-supervised & None \\
23 & Lu and Ye \cite{Lu21} & Supervised & Supervised & None \\
24 & Ma et al. \cite{Ma21} & None & Zero-shot & None \\
25 & Niu et al. \cite{Niu22} & None & Meta & None \\
26 & Ouyang et al. \cite{Ouyang22} & Supervised & Self-supervised and Meta & None \\
27 & Pham et al. \cite{Pham20} & None & Supervised & Classical \\
28 & Pham, Dovletov and Pauli \cite{Pham21} & None & Supervised & None \\
29 & Roy et al. \cite{Roy20} & None & Supervised & None \\
30 & Roychowdhury et al. \cite{Roychowdhury21} & None & Supervised & Classical \\
31 & Rutter, Lagergren and Flores \cite{Rutter19} & None & Semi-supervised & Classical \\
32 & Shen et al. \cite{Shen20} & None & Semi-supervised & Registration-based\\
33 & Shen et al. \cite{Shen21} & Supervised & Semi-supervised & None \\
34 & Shi et al. \cite{Shi23} & Supervised & Semi-supervised & Registration-based\\
35 & Sun et al. \cite{Sun22} & Supervised & Meta & None \\
36 & Tang et al. \cite{Tang21} & None & Meta & None \\
37 & Tomar et al. \cite{Tomar22} & Self-supervised & Supervised & Generative and Registration-based \\
38 & Wang et al. \cite{Wang20} & None & Semi-supervised & None\\
39 & Wang et al. \cite{Wang21} & None & Supervised & Classical \\
40 & Wang et al. \cite{Wang221} & None & Semi-supervised & Classical \\
\bottomrule
\end{tabular}
\end{table*}

\FloatBarrier
\begin{table*}[ht]
\ContinuedFloat
\caption{\emph{(continued)}.\label{tab:main_pipeline_table_seg}}
\small
\centering
\begin{tabular}
{P{.1\textwidth}P{.2\textwidth}P{.25\textwidth}P{.15\textwidth}P{.2\textwidth}}\\
\toprule
\textbf{Study ID} & \textbf{Pub. ref.} & \textbf{Pre-training} & \textbf{Training}&\textbf{Data augmentation}\\
\toprule
41 & Wang, Zhou and Zheng \cite{Wang322} & Supervised & Meta & None \\
42 & Wang et al. \cite{Wang22} & None & Supervised & None\\
43 & Wu, Xiao and Liang \cite{Wu22} & None & Supervised & Classical\\
44 & Wu et al. \cite{Wu222} & None & Semi-supervised & None \\
45 & Xu and Niethammer \cite{Xu19} & Supervised & Semi-supervised & Classical and Registration-based \\
46 & Yu et al. \cite{Yu21} & Supervised & Meta & Classical \\
47 & Yuan, Esteva and Xu \cite{Yuan21} & None & Supervised and Meta & Classical\\
48 & Zhao et al. \cite{Zhao19} & None & Semi-supervised & Generative and Registration-based \\
49 & Zhao et al. \cite{Zhao22} & None & Meta & Classical and Generative \\
50 & Zhou et al. \cite{Zhou21} & Supervised & Supervised & None \\
\bottomrule
\end{tabular}
\end{table*}
\FloatBarrier

\subsubsection{Classification}
\noindent We identified 27 relevant studies, each focusing on medical classification as its primary task. To enhance clarity and facilitate easy reference, we present all the information extracted from the selected studies in Table \ref{tab:FSL_classification}. In addition, we provide information concerning ROB and the applicability concerns in Table \ref{tab:Bias_FSL_classification}.

{\color{red} As for segmentation, different strategies are present in SOTA for addressing FSL classification in the medical imaging domain. Regarding the meta-learning techniques, for example, Dai et al. \cite{Dai23} introduced PFEMed, i.e., an episodic-trained model that leverages a dual-encoder structure and a prior-guided Variational Autoencoder (VAE) to enhance the robustness of the target features w.r.t. classify query samples. We observed that meta-training can also serve as a pre-training step before fine-tuning on the downstream task. This approach was adopted by Maicas et al. \cite{Maicas19}, who proposed to meta-train the model on a wide selection of tasks and fine-tuning the pre-trained model on the target task using a classical fully-supervised approach.
In terms of other learning techniques, Mahapatra et al. \cite{Mahapatra22} addressed GZSL by exploiting an SSL to generate anchor vectors for seen and unseen classes. Anchor vectors of the seen class samples were used to get SSL-based loss terms for the unseen class samples clustering stage. As a second step, they performed feature generation and used synthesized and actual features of unseen and seen classes to train a classifier.}

\FloatBarrier
\begin{table*}[ht]
\small
\caption{FSL studies for medical image classification.}\label{tab:FSL_classification}
\centering
\begin{tabular}[!t]
{P{.05\textwidth}P{.15\textwidth}P{.25\textwidth}P{.1\textwidth}P{.2\textwidth}P{.15\textwidth}}\\
\toprule
\textbf{Study ID} &\textbf{Pub. ref.} & \textbf{Algorithm/Pipeline} & \textbf{K-shot} & \textbf{Best performance} & \textbf{Meta-learning type}\\
\toprule
51 & Ali et al. \cite{Ali20}& \makecell{Prototypical \\network}	& 5-shot	& \makecell{Accuracy: \\ 0.906 \\(Endoscopic images,\\ MAO)} &Metric learning
\\
\midrule
52 & Cai, Hu, and Zheng \cite{Cai20}	&  \makecell{Prototypical \\ network + \\ Attention\\ module (CBAM) } &	20-shot	&	\makecell{Accuracy:\\ 0.924 (Brain, MAT)} & Metric learning
\\
\midrule
53 & Cai et al. \cite{Cai22} & \makecell{Pre-Moco\\ Diagnosis Network \\ (Pre-training+\\Contrastive learning)} & \makecell{1-shot \\ 5-shot\\ 10-shot \\ 20-shot} & \makecell{Accuracy: \\ 0.832 (Skin, MAD) \\
0.675 (Eye, MAD)} & Metric learning\\
\midrule
54 & Cano and Cruz-Roa \cite{Cano20}	& 	\makecell{Siamese \\ Neural Network}	& 1-shot	& \makecell{Accuracy: \\ 0.908 (Breast, MAT)} & Metric learning\\
\midrule
55 & Chen et al. \cite{Chen21}  &	\makecell{2D CNN ranking + \\2D CNN classification +\\ Heatmap for segmentation} & 2-shot	& \makecell{AUROC:\\ 0.883 (Breast,\\ LN metastases)}& None\\
\midrule
56 &Chou et al. \cite{Chou22} & 	\makecell{Siamese Neural Network\\ (Triple encoder +\\ Triple loss)}&1-shot & \makecell{Accuracy: \\ 0.986 (Brain,\\ classification\\ into contrast \\type)}& None\\
\bottomrule
\end{tabular}
\end{table*}
\FloatBarrier

\FloatBarrier
\begin{table*}[ht]
\ContinuedFloat
\small
\caption{\emph{(continued).}\label{tab:FSL_classification}}
\centering
\begin{tabular}[!t]
{P{.05\textwidth}P{.15\textwidth}P{.25\textwidth}P{.1\textwidth}P{.2\textwidth}P{.15\textwidth}}\\
\toprule
\textbf{Study ID} &\textbf{Pub. ref.} & \textbf{Algorithm/Pipeline} & \textbf{K-shot} & \textbf{Best performance} & \textbf{Meta-learning type} \\
\toprule
57 & Dai et al. \cite{Dai23} & \makecell{Prior Guided Feature \\Enhancement for \\ Few-shot Medical \\Image Classification} & \makecell{3-shot \\ 5-shot\\10-shot} & \makecell{Accuracy: \\ 0.851 (Brain, MAT) \\ 0.960 (Skin, MAT) \\ 0.803 (Cervix, MAT)} & Metric learning
\\
\midrule
58 & Huang, Huang and Tang \cite{Huang22} & \makecell{One-shot Anomaly \\Detection Framework} & 1-shot & \makecell{ AUROC: \\ 0.961 (Eye, MAD)\\ 0.955 (Lung ,COVID)} & None \\
\midrule
59 & Jiang et al. \cite{Jiang22} & \makecell{Autoencoder + \\Metric learner + \\Task learner (Transfer \\learning phase +\\ Meta-learning phase)} & \makecell{1-shot \\ 5-shot\\10-shot} & \makecell{Accuracy: \\0.762 (Cells, MAS) \\ 0.762 
(Colon, MAD)\\ 0.506 (Lungs, MAD) } & Metric learning
\\
\midrule
60 & Jin et al. \cite{Jin21} 	&	\makecell{ViT-L/16 +\\ ResNet50 +\\ Metric-learning }	& \makecell{1-shot\\5-shot\\8-shot} & \makecell{Accuracy: \\ 0.346 (Lungs, MAD)}	& Metric learning\\
\midrule
61 & Mahapatra, Ge and Reyes \cite{Mahapatra22}	& \makecell{Self-Supervised\\ Clustering Based\\ Generalized Zero-shot\\ Learning} &	0-shot	& \makecell{Accuracy: \\ 0.921 (Breast, \\LN metastases) \\ 0.909 (Lungs, MAD) \\ 0.942 (Eye, DE) \\ 0.911 (Prostate, tumour)}	& None
\\
\midrule
62 & Maicas et al. \cite{Maicas19}	& \makecell{Pre and post-hoc\\ diagnosis and \\interpretation +\\ 3D DenseNet}& 4-shot	& 	\makecell{AUROC: \\ 0,910 (Breast, tumour)}& Initialization\\
\midrule
63 & Mohan et al. \cite{Mohan21} 	& 	\makecell{Siamese Network +\\ Classifier} & 1-shot &	\makecell{Accuracy: \\ 0.930 \\(Lung,\\ COVID and\\ Pneumonia)}& None
\\
\midrule
64 & Moukheiber et al. \cite{Moukheiber22} & \makecell{DeepVoro\\ Multi-label \\ensemble} & \makecell{5-shot \\ 10-shot} & \makecell{AUROC: \\ 0.679 (Lung, MAD)} & Initialization-based\\
\midrule
65  & Naren, Zhu and Wang \cite{Naren21} 	& \makecell{8 block VGG + \\ MAML++}& \makecell{1-shot to \\ 5-shot}	& \makecell{Accuracy: \\ 0.857 (Lung, COVID)} &  Initialization\\
\midrule
66 & Ouahab, Ben-Ahmed and Fernandez-Maloigne \cite{Ouahab22} & \makecell{Self-attention \\augmented MAML} & \makecell{3-shot\\5-shot}	& \makecell{Accuracy: \\ 0.819 
 (Skin, MAD) \\ 0.703 (Lungs, MAD) \\ AUROC: \\ 0.843 (Skin,MAD) \\ 
 0.734 (Lungs, MAD)}	& Initialization\\
\midrule
67 & Paul, Tang and Summers \cite{Paul20} & \makecell{DenseNet-121\\ (feature extractor) +\\ Autoencoder ensemble \\(classificator)} &	5-shot	&	\makecell{F1-score:\\ 0.440 (Lung, MAD)\\ Recall: \\0.490 (Lung,MAD) }& None\\
\bottomrule
\end{tabular}
\end{table*}
\FloatBarrier

\FloatBarrier
\begin{table*}[ht]
\ContinuedFloat
\small
\caption{\emph{(continued).}\label{tab:FSL_classification}}
\centering
\begin{tabular}[!t]
{P{.05\textwidth}P{.15\textwidth}P{.25\textwidth}P{.1\textwidth}P{.2\textwidth}P{.15\textwidth}}\\
\toprule
\textbf{Study ID} &\textbf{Pub. ref.} & \textbf{Algorithm/Pipeline} & \textbf{K-shot} & \textbf{Best performance} & \textbf{Meta-learning type} \\
\toprule
68 & Paul et al. \cite{Paul321}	& \makecell{DenseNet + \\ Vanilla \\autoencoder )	}& 5-shot	&	\makecell{F1-score:\\ 0.470 (Lungs, MAD) \\  AUROC: \\0.647 (Lungs, MAD) }&	None\\
\midrule
69 & Paul et al. \cite{Paul221}	& \makecell{DenseNet +\\ MVSE network +\\ Self-training}	& 0-shot &	\makecell{Recall:\\ 0.454 (Lungs, MAD)}& None\\
\midrule
30 & Roychowdhury et al. \cite{Roychowdhury21}	& \makecell{Echo state network \\(ParESN) +\\ Target label\\ selection algorithm\\ (TLSA) } &5-shot &\makecell{Accuracy:\\ 0.970 (Eye, IE)
}& None\\
\midrule
70 & Singh et al. \cite{Singh21}	& \makecell{MetaMed}& \makecell{3-shot\\5-shot\\10-shot}&	\makecell{Accuracy:\\ 0.864 (Breast, MAD) \\ 0.843 (Skin, MAD) \\ 0.934
 (Cervix, MAT) }& Initialization\\
\midrule
71 & Vetil et al. \cite{Vetil22}	& \makecell{VAE + \\Distribution\\ learning} &	\makecell{0-shot\\ 15-shot}	& \makecell{AUROC: \\ 0.789 (Pancreas)}& None \\
\midrule
72 & Xiao et al. \cite{Xiao21}& \makecell{CNN \\feature extractor +\\ classification\\ prototype + \\similarity\\ module +\\ rectified \\corruption function}	& \makecell{5-shot\\10-shot}& \makecell{Accuracy:\\ 0.874 (Skin, MAD)}& Metric learning\\
\midrule
73 & Yan et al. \cite{Yan22}& \makecell{Siamese-\\Prototypical Network} &	\makecell{1-shot \\ 5-shot}	&	\makecell{Accuracy: \\0.686 (Skin, MAD) \\0.608
 (Liver, MAD) \\ 0.626 (Colon, MAD)}& Metric learning\\
\midrule
74 &  Yarlagadda et al. \cite{Yarlagadda19}	& \makecell{Region proposal network + \\Inception-ResNet-v2 +\\ Memory module with \\regional maximum activation\\ of convolutions global\\ descriptors} & 1-shot	& \makecell{Accuracy: \\0.946 (Cells)}	& None
\\
\midrule
75 & Zhang, Cui and Ren \cite{Zhang22} &  MAML & \makecell{1-shot\\ 3-shot\\5-shot}	& \makecell{Accuracy: \\0.788 (VQA-\\RAD, MAS) \\ 0.614 (PathVQA,\\MAS)}&	Initialization
\\
\midrule
76 & Zhu et al. \cite{Zhu20}	& \makecell{Query-Relative \\Loss + Adaptive \\Hard Margin +\\ Prototypical Network/\\  Matching Network} & \makecell{1-shot \\5-shot}	& \makecell{Accuracy:\\ 0.719 (Skin, MAD)}
 & Metric learning
\\
\bottomrule
\end{tabular}
\end{table*}
\FloatBarrier

\FloatBarrier
\begin{table*}[ht]
\small
\caption{ROB of FSL studies for medical image classification.}\label{tab:Bias_FSL_classification}
\centering
\begin{tabular}
{P{.05\textwidth}P{.2\textwidth}P{.05\textwidth}P{.05\textwidth}P{.05\textwidth}P{.08\textwidth}P{.08\textwidth}P{.05\textwidth}P{.05\textwidth}P{.05\textwidth}P{.05\textwidth}}\\
\toprule
&& \multicolumn{5}{c}{\textbf{Risk of Bias}} & \multicolumn{4}{c}{\textbf{Applicability}}\\
\cmidrule(lr){3-7}
\cmidrule(lr){8-11}
\textbf{Study ID}&	\textbf{Pub. ref.} &\textbf{Part.}& \textbf{Pred.} & \textbf{Out.} & \textbf{Analysis} & \textbf{Overall} & \textbf{Part.} & \textbf{Pred.} & \textbf{Out.} & \textbf{Overall}\\
\toprule
51	& Ali et al. \cite{Ali20}&	\greencheck&	\greencheck&	\greencheck	&\redcross&	\redcross&	\greencheck&	\greencheck&	\greencheck&	\greencheck\\
\midrule
52	& Cai, Hu, and Zheng \cite{Cai20}&	\greencheck&	\greencheck&	\greencheck&	\greencheck&	\greencheck&	\greencheck&	\greencheck&	\greencheck&	\greencheck\\
\midrule
53	& Cai et al. \cite{Cai22} &	\greencheck&	\greencheck&	\greencheck&	\greencheck&	\greencheck&	\greencheck&	\greencheck&	\greencheck&	\greencheck\\
\midrule
54	& Cano and Cruz-Roa \cite{Cano20}&	\greencheck&	\greencheck&	\greencheck&	\redcross	&\redcross&	\greencheck&	\greencheck&	\greencheck&	\greencheck\\   
\midrule
55	&Chen et al. \cite{Chen19}&	\greencheck&	\greencheck&	\greencheck&	\redcross	&\redcross&	\greencheck&	\greencheck&	\greencheck&	\greencheck\\ 
\midrule
56	&Chou et al. \cite{Chou22}	& \greencheck&	\greencheck	&\greencheck&	\redcross	&\redcross&	\greencheck&	\greencheck&	\greencheck	&\greencheck\\
\midrule
57	& Dai et al. \cite{Dai23}&	\greencheck&	\greencheck&	\greencheck&	\greencheck&	\greencheck&	\greencheck&	\greencheck&	\greencheck&	\greencheck  \\
\midrule
58	& Huang, Huang and Tang \cite{Huang22}&	\greencheck	&\greencheck	&\greencheck&	\greencheck&	\greencheck&	\greencheck&	\greencheck&	\greencheck&	\greencheck\\
\midrule
59	& Jiang et al. \cite{Jiang22}	&\greencheck	&\greencheck&	\greencheck	&\greencheck&	\greencheck&	\greencheck&	\greencheck&	\greencheck&	\greencheck\\
\midrule
60	& Jin et al. \cite{Jin21}	& \greencheck	& \greencheck&	\greencheck	& \greencheck	&\greencheck&	\greencheck&	\greencheck&	\greencheck&	\greencheck\\ 
\midrule
61	& Mahapatra, Ge and Reyes \cite{Mahapatra22}	& \greencheck&	\greencheck&	\greencheck&	\greencheck&	\greencheck	&\greencheck&	\greencheck&	\greencheck	&\greencheck\\
\midrule
62	& Maicas et al. \cite{Maicas19}	&\greencheck	&\greencheck&	\greencheck	&\greencheck&	\greencheck	&\greencheck&	\greencheck	&\greencheck&	\greencheck\\
\midrule
63	& Mohan et al. \cite{Mohan21}	&\greencheck	&\greencheck&	\greencheck	&\redcross&	\redcross&	\greencheck&	\greencheck&	\greencheck&	\greencheck\\
\midrule
64	& Moukheiber et al. \cite{Moukheiber22}&	\greencheck&	\greencheck&	\greencheck&	\greencheck	&\greencheck	&\greencheck&	\greencheck&	\greencheck&	\greencheck\\
\midrule
65	& Naren, Zhu and Wang \cite{Naren21}& 	\greencheck&	\greencheck&	\greencheck&	\greencheck&	\greencheck&	\greencheck&	\greencheck&	\greencheck&	\greencheck\\
\midrule
66	&Ouahab, Ben-Ahmed and Fernandez-Maloigne \cite{Ouahab22}&	\greencheck	&\greencheck&	\greencheck&	\greencheck	&\greencheck&	\greencheck&	\greencheck&	\greencheck&	\greencheck\\
\midrule
67	& Paul, Tang and Summers \cite{Paul20}	& \greencheck	& \greencheck&	\greencheck&	\greencheck&	\greencheck	&\greencheck&	\greencheck&	\greencheck	&\greencheck\\
68	&Paul et al. \cite{Paul321}	&\greencheck	&\greencheck&	\greencheck&	\greencheck&	\greencheck&	\greencheck&	\greencheck&	\greencheck&	\greencheck\\
\midrule
69	&Paul et al. \cite{Paul221}	&\greencheck	&\greencheck &	\greencheck&	\greencheck&	\greencheck	&\greencheck&	\greencheck&	\greencheck&	\greencheck\\
\midrule
30	&Roychowdhury et al. \cite{Roychowdhury21}	&\greencheck&	\greencheck&	\greencheck&	\redcross&	\redcross&	\greencheck&	\greencheck&	\greencheck&	\greencheck\\
\midrule
70	&Singh et al. \cite{Singh21}&	\greencheck&	\greencheck&	\greencheck&	\greencheck&	\greencheck&	\greencheck&	\greencheck&	\greencheck&	\greencheck\\
\midrule
71	&Vetil et al. \cite{Vetil22}	&\greencheck&	\greencheck&	\greencheck&	\greencheck&	\greencheck&	\greencheck&	\greencheck&	\greencheck&	\greencheck\\
\midrule
72	&Xiao et al. \cite{Xiao21}&	\greencheck	&\greencheck&	\greencheck&	\greencheck&	\greencheck&	\greencheck&	\greencheck&	\greencheck&	\greencheck\\
\midrule
73	&Yan et al. \cite{Yan22}	&\greencheck	&\greencheck	&\greencheck	&\greencheck	&\greencheck	&\greencheck	&\greencheck	&\greencheck	&\greencheck\\
\midrule
74	&Yarlagadda et al. \cite{Yarlagadda19}	&\greencheck	&\greencheck&	\greencheck&	\redcross	&\redcross	&\greencheck&	\greencheck	&\greencheck	&\greencheck\\
\midrule
75	&Zhang, Cui and Ren \cite{Zhang22}	&\greencheck	&\greencheck	&\greencheck	&\redcross	&\redcross	&\greencheck	&\greencheck	&\greencheck	&\greencheck\\
\midrule
76	&Zhu et al. \cite{Zhu20}&	\greencheck	&\greencheck	&\greencheck	&\greencheck	&\greencheck	&\greencheck	&\greencheck	&\greencheck	&\greencheck\\
\bottomrule
\end{tabular}
\end{table*}
\FloatBarrier

\noindent {\color{red} Here, we present the findings of our classification papers analysis.}

\begin{figure*}[ht]
\centering
\subfloat[Grouped by clinical task]{\includegraphics[width=0.5\textwidth]{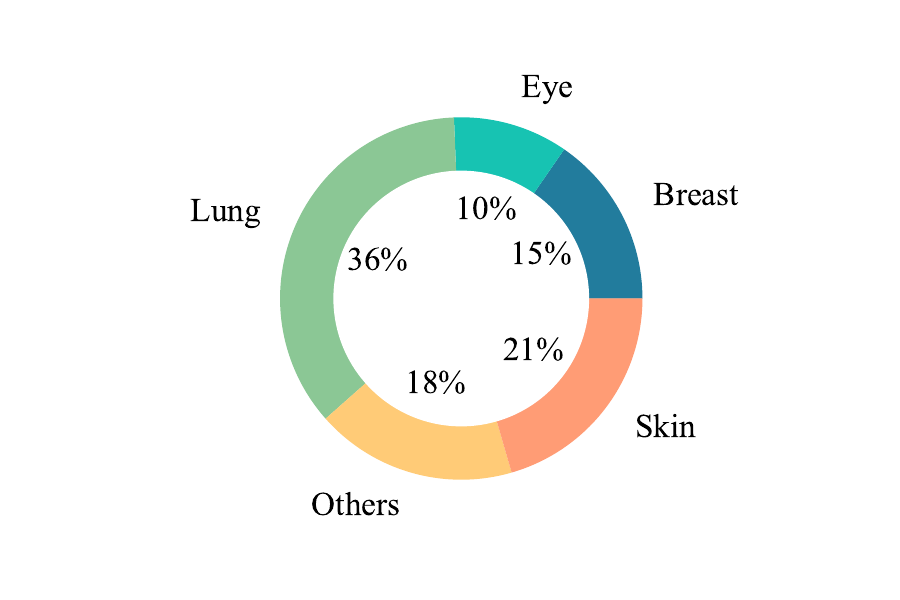}
\label{fig:chart_cls_organ}}
\subfloat[Grouped by meta-learning method]{\includegraphics[width=0.5\textwidth]{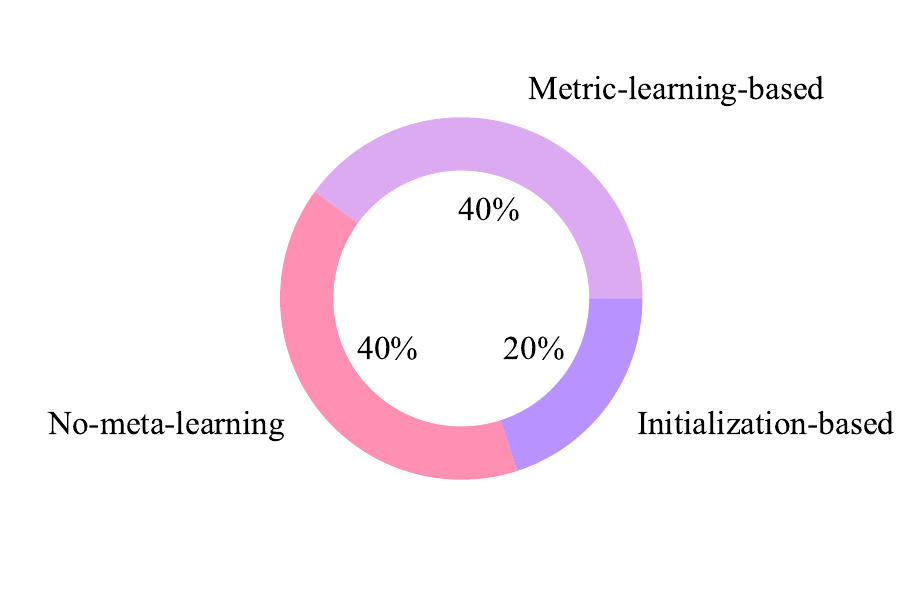}
\label{fig:chart_cls_method}}
\caption{Classification studies distribution grouped by clinical task and meta-learning method.}
\label{fig:charts_cls}
\end{figure*}

\noindent\textbf{\color{red}{Clinical task.}}
The classification papers covered a wide range of anatomical structures and regions, as well as various diseases. In the following, we provide a breakdown of articles categorized by the clinical task investigated: two out of 27 studies (7\%) performed brain image classification, focusing on different types of tumours and MRI contrast types; six studies (22\%) addressed breast image classification, with four concentrating on breast tumours and two on breast metastases involving nearby lymph nodes; two studies (7\%) investigated cell image classification; two studies (7\%) focused on cervix image classification; three studies (11\%) pertained to colon image classification; four studies (15\%) were dedicated to fundus eye image classification, with 2 investigating different diseases; one study (4\%) dealt with liver disease classification; 11 studies (41\%) involved lung image classification; one study (4\%) concerned  pancreas image classification; one study (4\%) classified prostate tumour images; 7 studies (26\%) addressed skin image classification, covering different diseases; one study (4\%) investigated esophagus image classification; one study (4\%) focused on stomach image classification.
Note that \cite{Zhang22} is not included in this analysis as it did not specify which anatomical structures were part of their study. An illustration of the classification studies distribution in terms of the five most popular clinical task investigated is provided in Figure \ref{fig:chart_cls_organ}.

\noindent\textbf{\color{red} {Meta-learning method.}}
In the context of classification studies employing FSL, the distribution of meta-learning methods is as follows: six out of 27 studies (22\%) utilized initialization-based methods; 10 studies (37\%) opted for metric-learning-based algorithms; the remaining 11 studies (41\%) did not incorporate any meta-learning technique.
A representation of the classification studies distribution in terms of the meta-learning methods employed is provided in Figure \ref{fig:chart_cls_method}.

\noindent\textbf{K-shot.}
Among the 27 selected studies in the classification domain using FSL, the training configurations were distributed as follows: 13 studies (48\%) employed k-shot training with k ranging from 2 to 20; six studies (22\%) utilized both OSL and FSL; one study (4\%) used both FSL and ZSL; five studies (19\%) exclusively performed OSL and two studies (7\%) solely employed ZSL.

\noindent\textbf{{\color{red} Image modality.}}
{\color{red} In the following, we provide the distribution of studies in terms of the imaging modalities utilized:} three studies out of 27 (11\%) used CT images; three studies (11\%) employed MRI images; seven studies (26\%) utilized dermoscopic images; 11 studies (41\%) relied on X-ray images; three studies (11\%) involved fundus images; two studies (7\%) made use of microscopic images; nine studies (33\%) investigated histopathological images; one study (4\%) utilized endoscopy images; one study (4\%) involved cytological images and one study (4\%) used OCT images.

\noindent\textbf{Model evaluation.} For assessing the robustness of the models, various evaluation techniques were employed: nine studies (33\%) utilized ablation studies; one study (4\%) conducted both ablation studies and cross-validation; one study (4\%) solely relied on cross-validation; two studies (7\%) repeated experiments multiple times for evaluation. The remaining 14 studies (52\%) did not employ any specific model evaluation technique.

\begin{figure*}[h!]
    \centering
    \includegraphics[width=13cm]{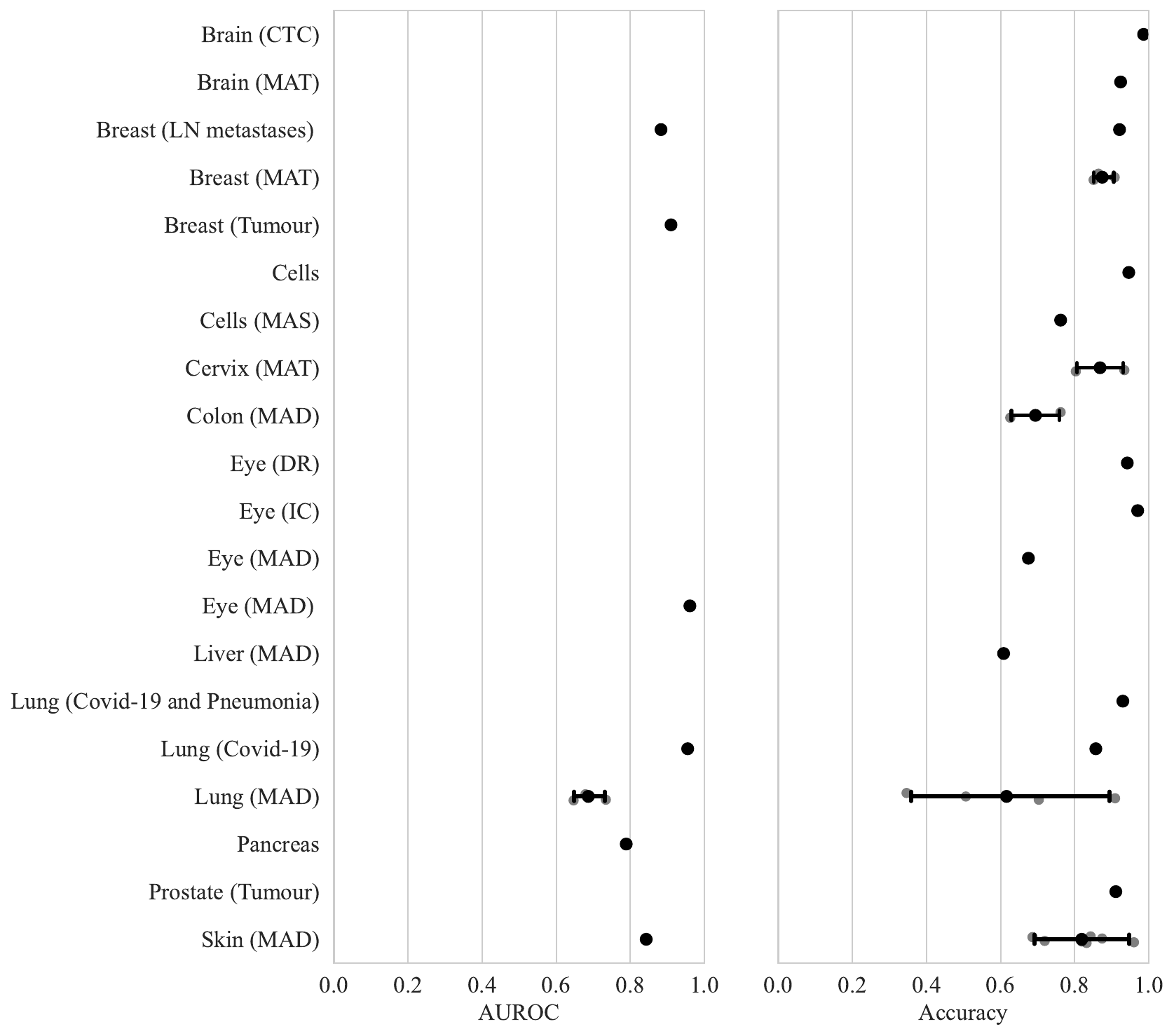}
    \caption{Forest plot of classification studies performance based on AUROC and Accuracy metrics. Studies are grouped by the clinical task investigated. CTC = Contrast-type Classification;  DR = Diabetic Retinopathy; IC = Intraretinal Cyst; LN = Limph Nodes; MAD = Mead Across Diseases; MAS = Mean Across Structures; MAT = Mean Across Tumours.}
    \label{fig:fp_cls_organ}
\end{figure*}

\begin{figure*}[h!]
    \centering
    \includegraphics[width=15cm]{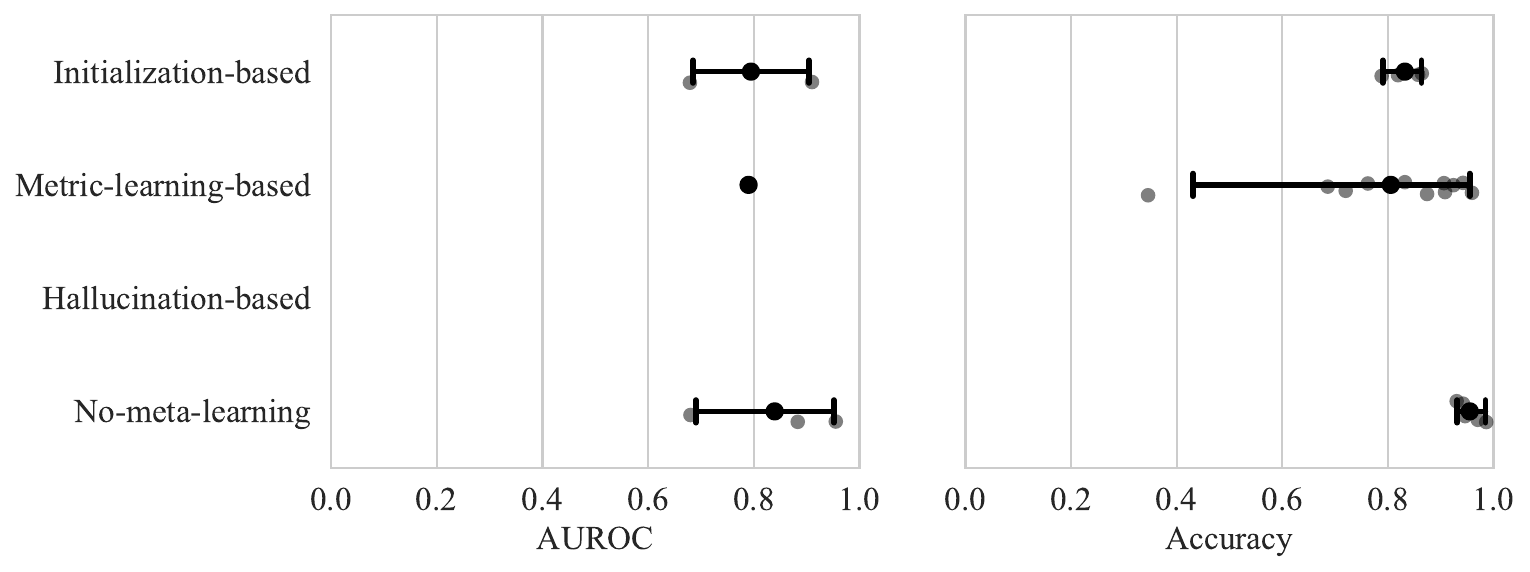}
    \caption{Forest plot of classification studies performance based on AUROC and Accuracy metrics. Studies are grouped by the meta-learning method employed.}
    \label{fig:fp_cls_method}
\end{figure*}

\noindent\textbf{{\color{red} Statistical analysis.}} {\color{red} Figure~\ref{fig:fp_cls_organ} displays forest plots that summarize the performance of FSL models (mean and 95\% CI) for each clinical task considering both AUROC score and accuracy metrics.
Conversely, Figure~\ref{fig:fp_cls_method} depicts the models' results grouped by the meta-learning method used. Note that the results of \cite{Ali20} and \cite{Zhang22} were not included in the forest plots since they provide average results across different clinical tasks.}

\noindent \textbf{{\color{red} Standard pipeline.}} {\color{red} Table~\ref{tab:main_pipeline_table_cls} outlines the learning frameworks employed in each step of the standard pipeline by the classification studies analyzed in this review. In the following, we explore the distribution of the studies based on the techniques employed for each pipeline phase:} one out of 27 studies (4\%) employed a meta-learning algorithm for pre-training; 13 studies (48\%) employed classical supervised pre-training; one study (4\%) used unsupervised pre-training; 12 studies (44\%) did not employ any pre-training stage. For training, fifteen studies (56\%) utilized meta-learning; one study (4\%) employed semi-supervised training; one study (4\%) employed self-supervised training; nine studies (33\%) used traditional supervised training; two studies (7\%) employed zero-shot learning methods. Finally, concerning the data augmentation techniques, 10 out of 27 studies (37\%) relied on classical data augmentation techniques; two studies (7\%) utilized generative methods for data augmentation. The remaining 15 studies did not employ data augmentation.

\FloatBarrier
\begin{table*}[ht]
\small
\caption{Standard pipeline steps adopted by classification studies.\label{tab:main_pipeline_table_cls}}
\centering
\begin{tabular}[!t]
{P{.1\textwidth}P{.2\textwidth}P{.25\textwidth}P{.15\textwidth}P{.2\textwidth}}\\
\toprule
\textbf{Study ID} & \textbf{Pub. ref.} & \textbf{Pre-training} & \textbf{Training}&\textbf{Data augmentation}\\
\toprule
51 & Ali et al. \cite{Ali20} & Supervised & Meta & None \\
52 & Cai, Hu, and Zheng \cite{Cai20} & None & Meta & Classical \\
53 & Cai et al. \cite{Cai22} & Supervised & Meta & Classical \\
54 & Cano and Cruz-Roa \cite{Cano20} & None & Meta & None \\
55 & Chen et al. \cite{Chen21} & Unsupervised & Supervised & None \\
56 & Chou et al. \cite{Chou22} & None & Supervised & None \\
57 & Dai et al. \cite{Dai23} & Supervised & Meta & None \\
58 & Huang, Huang and Tang \cite{Huang22} & None & Supervised & Generative \\
59 & Jiang et al. \cite{Jiang22} & Supervised & Meta & Classical \\
60 & Jin et al. \cite{Jin21} & None & Meta & Classical \\
61 & Mahapatra, Ge and Reyes \cite{Mahapatra22} & Supervised & Zero-shot  Self-supervised and Supervised & Generative \\
62 & Maicas et al. \cite{Maicas19} & Meta & Supervised & None \\
63 & Mohan et al. \cite{Mohan21} & Supervised & Supervised & Classical \\
64 & Moukheiber et al. \cite{Moukheiber22} & Supervised & Meta & None \\
65 & Naren, Zhu and Wang \cite{Naren21} & None & Meta & None \\
66 & Ouahab, Ben-Ahmed and Fernandez-Maloigne \cite{Ouahab22} & Supervised & Meta & Classical \\
67 & Paul, Tang and Summers \cite{Paul20} & Supervised & Supervised & None\\
68 & Paul et al. \cite{Paul221} & Supervised & Zero-shot and Semi-supervised & None \\
69 & Paul et al. \cite{Paul321} & Supervised & Supervised & None \\
30 & Roychowdhury et al. \cite{Roychowdhury21} & None & Supervised & Classical \\
70 & Singh et al. \cite{Singh21} & None & Meta & Classical \\
71 & Vétil et al. \cite{Vetil22} & None &  Zero-shot and Supervised & Classical \\
72 & Xiao et al. \cite{Xiao21} & None & Meta & None \\
73 & Yan et al. \cite{Yan22} & Supervised & Meta & Classical \\
74 & Yarlagadda et al. \cite{Yarlagadda19} & Supervised & Supervised & None \\
75 & Zhang, Cui and Ren \cite{Zhang22} & None & Meta & None \\
76 & Zhu et al. \cite{Zhu20} & None & Meta & None \\
\bottomrule
\end{tabular}
\end{table*}
\FloatBarrier

\subsubsection{Registration}
\noindent We included six relevant studies, each focusing on medical registration as its primary task. Table \ref{tab:FSL_registration} summarizes all the information extracted from the selected studies. In addition, we provide information concerning ROB and the applicability of each study in Table \ref{tab:Bias_FSL_registration}. {\color{red}Unlike segmentation and classification, medical image registration approaches for FSL generally do not employ meta-learning techniques but typically rely on an unsupervised learning framework where the model is trained until convergence using only the pair of images to be registered. An example of this is provided by Fechter et al. \cite{Fechter20}, who proposed a OSL registration method for periodic motion tracking that combines a U-Net architecture trained in a supervised way with a coarse-to-fine approach and a differential spatial transformer module. 
Two other works, instead, leveraged joint training between segmentation and registration networks and provided weak supervision to the registration network to improve performance. For example, Shi et al. \cite{Shi23} employed a segmentation module to predict pseudo-labels on unlabeled data to provide weak supervision for the registration network.}

\FloatBarrier
\begin{table*}[ht]
\small
\caption{FSL studies for medical image registration.}\label{tab:FSL_registration}
\centering
\begin{tabular}[!t]
{P{.05\textwidth}P{.15\textwidth}P{.25\textwidth}P{.1\textwidth}P{.2\textwidth}P{.15\textwidth}}\\
\toprule
\textbf{Study ID} &\textbf{Pub. ref.} & \textbf{Algorithm/Pipeline} & \textbf{K-shot} & \textbf{Best performance} & \textbf{Meta-learning type} \\
\toprule
77 & Fechter and Baltas \cite{Fechter20}	& \makecell{U-net + \\ Differential spatial \\transformer module} & \makecell{1-shot} & \makecell{ALD: \\ 1.49 (Lungs) \\ Dice: \\ 0.860 (Heart)} &None\\
\midrule
78 & Ferrante et al. \cite{Ferrante18}&	\makecell{U-net + \\Unsupervised \\learning} & 1-shot	& \makecell{Dice: \\ 0.920 (Heart)\\ 0.890 (Lungs)}& None\\
\midrule
79 & He et al. \cite{He222} & 	\makecell{Perception-\\Correspondence \\ Registration}	& 5-shot	& 	\makecell{Dice:\\ 0.857 (Heart, MAS) \\ 0.867 (Cervical \\vertebra, MAS) \\ 0.800 (Brain, MAS)}&	None\\
\midrule
34 & Shi et al. \cite{Shi23}& \makecell{Joint Registration \\and Segmentation\\ Self-training Framework} & 5-shot & \makecell{Dice: \\0.759 
 (Brain, MAS) \\ 0.539 (Abdomen, MAS)} & None \\
\midrule
45 & Xu and Niethammer \cite{Xu19} & \makecell{Semi-Supervised\\ Learning + \\Segmentation network + \\Registration
network} & \makecell{1-shot \\ 5-shot\\10-shot} & \makecell{Dice:\\ 0.759 (Brain, MAS)\\ 0.539
Abdomen (MAS)} & None
 \\
\midrule
80 & Zhang et al. \cite{Zhang221}& \makecell{CNN + \\Spatial transformer +\\ similarity loss +\\ smooth loss +\\ cyclic loss}	& 1-shot & \makecell{TRE:\\ 1.03 (Lung)}& 	None\\
\bottomrule
\end{tabular}
\end{table*}
\FloatBarrier

\FloatBarrier
\begin{table*}[ht]
\small
\caption{ROB of FSL studies for medical image registration.}\label{tab:Bias_FSL_registration}
\centering
\begin{tabular}
{P{.05\textwidth}P{.2\textwidth}P{.05\textwidth}P{.05\textwidth}P{.05\textwidth}P{.08\textwidth}P{.08\textwidth}P{.05\textwidth}P{.05\textwidth}P{.05\textwidth}P{.05\textwidth}}\\
\toprule
&& \multicolumn{5}{c}{\textbf{Risk of Bias}} & \multicolumn{4}{c}{\textbf{Applicability}}\\
\cmidrule(lr){3-7}
\cmidrule(lr){8-11}

\textbf{Study ID}&	\textbf{Pub. ref.} &\textbf{Part.}& \textbf{Pred} & \textbf{Out.} & \textbf{Analysis} & \textbf{Overall} & \textbf{Part.} & \textbf{Pred.} & \textbf{Out.} & \textbf{Overall}\\
\toprule
77	&Fechter, Baltas \cite{Fechter20}&	\greencheck&	\greencheck&	\greencheck&	\greencheck&	\greencheck	&\greencheck&	\greencheck&	\greencheck&	\greencheck\\
\midrule
78	&Ferrante et al. \cite{Ferrante18}.&	\greencheck&	\greencheck&	\greencheck&	\greencheck&	\greencheck&	\greencheck&	\greencheck&	\greencheck&	\greencheck\\
\midrule
79 &	He et al. \cite{He222}	&\greencheck	&\greencheck&	\greencheck&	\greencheck&	\greencheck&	\greencheck&	\greencheck&	\greencheck&	\greencheck\\
\midrule
34&	Shi et al. \cite{Shi23}	&\greencheck	&\greencheck	&\greencheck	&\greencheck	&\greencheck	&\greencheck	&\greencheck	&\greencheck	&\greencheck\\
\midrule
45	&Xu and Niethammer \cite{Xu19}&	\greencheck	&\greencheck&	\greencheck&	\greencheck&	\greencheck	&\greencheck&	\greencheck&	\greencheck&	\greencheck\\
\midrule
80	&Zhang et al. \cite{Zhang221}	&\greencheck&	\greencheck&	\greencheck&	\greencheck&	\greencheck&	\greencheck&	\greencheck&	\greencheck&	\greencheck\\
\bottomrule
\end{tabular}
\end{table*}
\FloatBarrier

\noindent Here, we present the findings derived from our comprehensive analysis of the registration papers.

\begin{figure*}[ht]
\centering
\subfloat[Grouped by clinical task]{\includegraphics[width=0.5\textwidth]{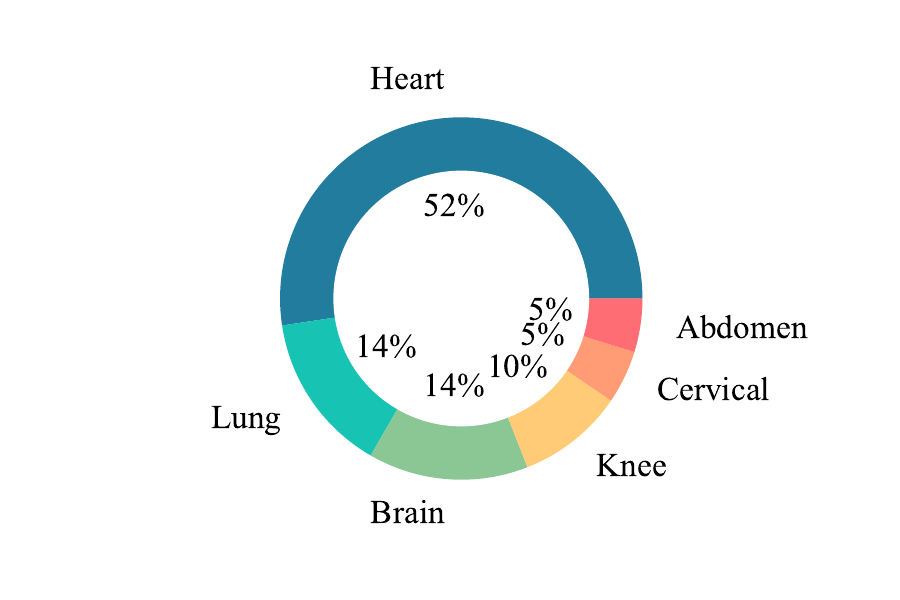}
\label{fig:chart_reg_organ}}
\subfloat[Grouped by meta-learning method]{\includegraphics[width=0.5\textwidth]{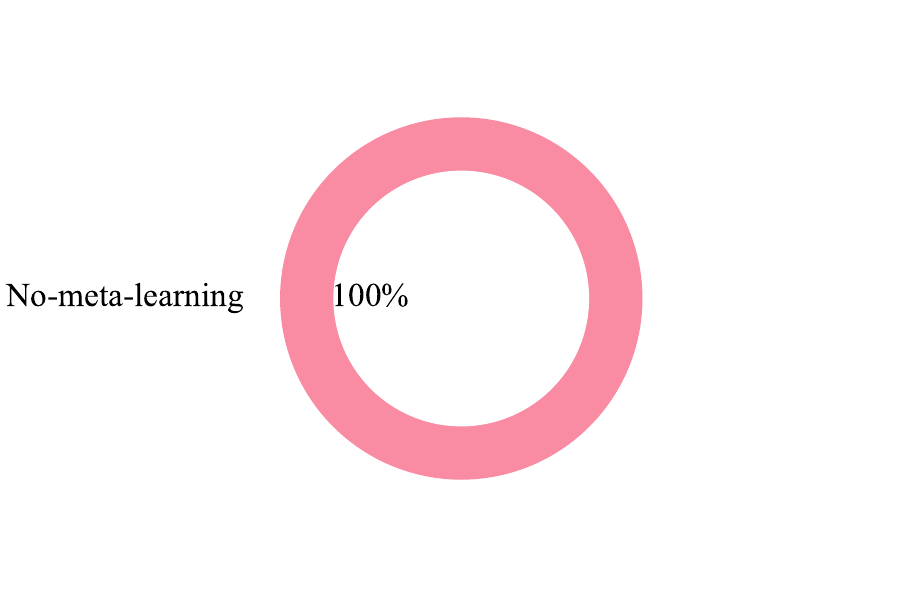}
\label{fig:chart_reg_method}}
\caption{Registration studies distribution grouped by clinical task and meta-learning method.}
\label{fig:charts_reg}
\end{figure*}

\noindent\textbf{{\color{red} Clinical task.}}
The selected registration papers investigate a range of anatomical regions. Here's the breakdown of studies w.r.t. the clinical task addressed: three out of 6 studies (50\%) explored brain registration; one study (17\%) focused on the registration of knee bones and cartilages; three studies (50\%) delved into heart registration; three studies (50\%) concentrated on lung registration; one study (17\%) pertained to abdominal registration; one study (17\%) dealt with cervical vertebra registration.
We represent the registration studies distribution w.r.t. clinical task in Figure \ref{fig:chart_reg_organ}.

\noindent\textbf{{\color{red} Meta-learning method.}}
In the registration studies domain, all of the selected papers (100 \%) did not employ the meta-learning paradigm, as illustrated in Figure \ref{fig:chart_reg_method}.

\noindent\textbf{K-shot.}
Among the six selected studies, the distribution of k-shot configurations is as follows: two studies (33\%) solely employed FSL; three studies (50\%) investigated solely OSL; one study (17\%) performed both FSL and OSL.

\noindent\textbf{{\color{red} Image modality.}}
In the context of registration studies, the distribution of imaging modalities used among the selected papers is as follows: four out of six studies (67\%) employ CT acquisitions; five out of six studies (83\%) utilize MRI images; one out of six studies (17\%) involves X-ray images.

\noindent\textbf{Model evaluation.}
To investigate the models' robustness, several evaluation techniques were employed: two studies (33\%) utilized only ablation studies and one study (17\%) used cross-validation. The remaining studies (50\%) did not employ any specific model evaluation technique.

\begin{figure*}[h!]
    \centering
    \includegraphics[width=13cm]{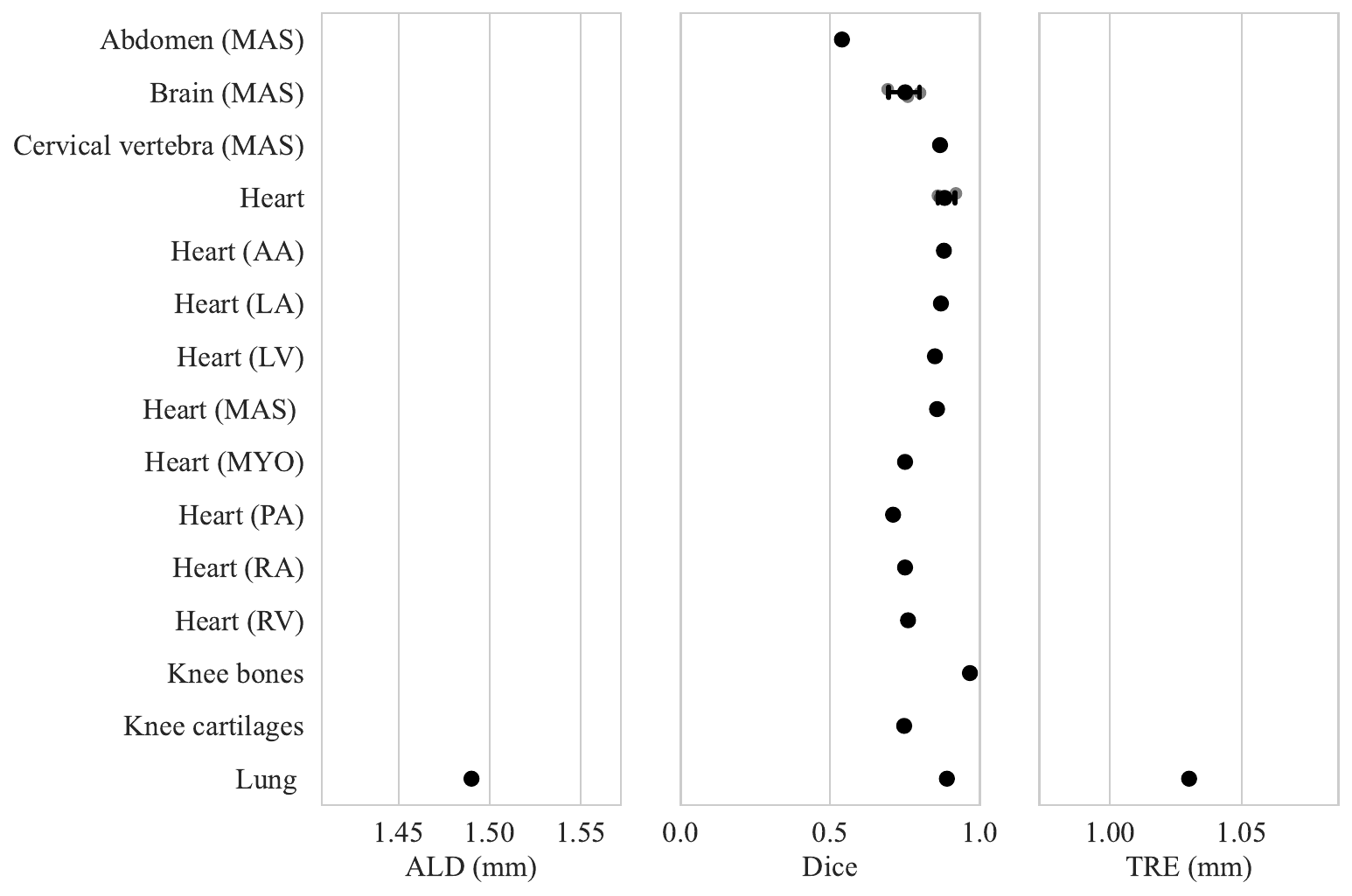}
    \caption{Forest plot of registration studies performance based on ALD, Dice, and TRE metrics. Studies are grouped by the clinical task investigated. AA = Ascending Aorta; LA = Left Atrium; LV = Left Ventricle; MAS = Mean Across Structures; MYO = Myocardium; PA = Pulmonary Artery; RA = Right Atrium; RV = Right Ventricle.}
    \label{fig:fp_reg_organ}
\end{figure*}

\begin{figure*}[h!]
    \centering
    \includegraphics[width=13cm]{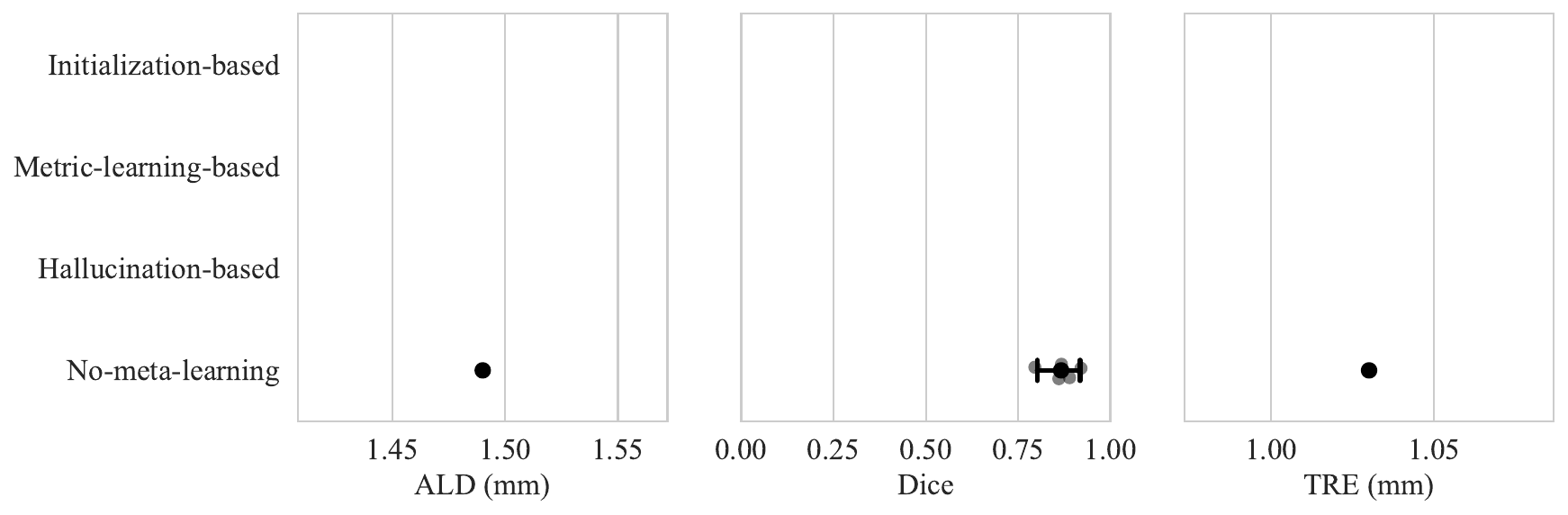}
    \caption{Forest plot of registration studies performance based on ALD, Dice and TRE metrics. Studies are grouped by the meta-learning method employed.}
    \label{fig:fp_reg_method}
\end{figure*}

\noindent\textbf{{\color{red} Statistical analysis.}}
{\color{red} \color{red} Figure~\ref{fig:fp_reg_organ} displays forest plots that summarize the performance of FSL models (mean and 95\% CI) for each clinical task considering ALD, Dice score and TRE metrics.
Conversely, Figure~\ref{fig:fp_reg_method} depicts the models' results grouped by the meta-learning method used. }

\noindent\textbf{{\color{red} Standard pipeline.}} {\color{red} Table~\ref{tab:main_pipeline_table_reg} outlines the learning frameworks employed in each step of the standard pipeline by the registration studies analyzed in this review. In the following, we explore the distribution of the studies based on the techniques employed for each pipeline phase:} two out of six studies (33\%) utilized a classical supervised approach in both pre-training and training steps. The remaining four studies (67\%) did not employ any pre-training and adopted an unsupervised approach during training. Two out of six studies (33\%) used classical data augmentation techniques. The other four studies (67\%) did not exploit data augmentation.

\FloatBarrier
\begin{table*}[ht]
\small
\caption{Standard pipeline steps adopted by registration studies.\label{tab:main_pipeline_table_reg}}
\centering
\begin{tabular}[!t]
{P{.1\textwidth}P{.2\textwidth}P{.25\textwidth}P{.15\textwidth}P{.2\textwidth}}\\
\toprule
\textbf{Study ID} & \textbf{Pub. ref.} & \textbf{Pre-training} & \textbf{Training}&\textbf{Data augmentation}\\
\toprule
77 & Fechter, Baltas \cite{Fechter20} & None & Unsupervised & None\\
78 & Ferrante et al. \cite{Ferrante18} & None & Unsupervised & None \\
79 & He et al. \cite{He222} & None & Unsupervised & Classical \\
34 & Shi et al. \cite{Shi23} & Supervised & Supervised & None\\
45 & Xu and Niethammer \cite{Xu19} & Supervised & Supervised & Classical \\
80 & Zhang et al. \cite{Zhang221} &  None & Unsupervised & None\\
\bottomrule
\end{tabular}
\end{table*}
\FloatBarrier

\subsection{Standard pipeline}

\begin{figure*}[h!]
    \centering
    \includegraphics[width=13cm]{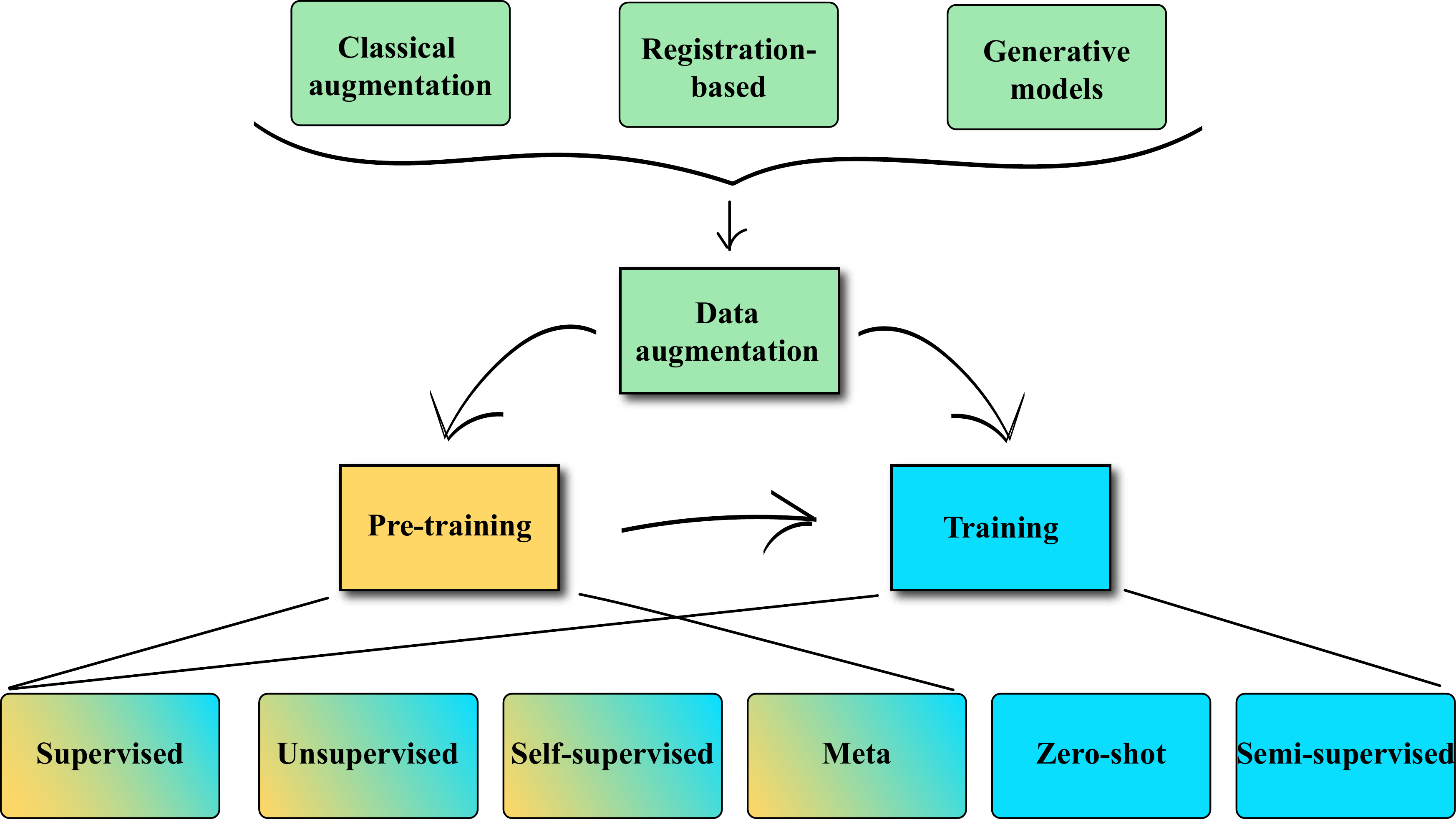}
    \caption{{\color{red} An illustration of the proposed standard pipeline. Learning frameworks suitable for both pre-training and training phases are highlighted in shaded boxes. For optimal clarity, please view this figure in colour.}}
    \label{fig:standard_pipeline}
\end{figure*}

\noindent {\color{red} Figure~\ref{fig:standard_pipeline} presents the proposed standardized pipeline for FSL in medical imaging. This pipeline was derived from a comprehensive analysis of all the selected studies across classification, segmentation, and registration tasks.}

\section{Discussion}
\label{sec:discussion}

\noindent {\color{red} This review analyzed 80 FSL studies applied to the medical imaging domain.} We organized the studies into three main categories based on their primary outcome: segmentation, classification, and registration.
{\color{red} Within each category, we extracted relevant information from each study, including the utilized algorithm or pipeline, employed meta-learning methods, the amount of labelled training data, and the highest performance achieved.
Additionally, we conducted a ROB and applicability analysis using the PROBAST method for each study. We further performed a statistical analysis, grouping the studies by both the investigated clinical task and the employed meta-learning techniques.
Furthermore, we analyzed the learning frameworks used in each study across three key phases: pre-training, training, and data augmentation. Based on this comprehensive analysis, we defined a general pipeline encompassing the techniques shared across the studies and relevant to the three targeted outcomes.}

{\color{red} In the following paragraphs, we discuss the results of our analysis based on the objectives outline provided in Section \ref{sec:objectives}. 
First, we examine the distribution of studies across various outcomes. Next, we discuss the results of our statistical analysis as well as additional aspects like data usage, imaging modalities, and model robustness assessment methods. Later, we provide insights into our main finding, i.e. the definition of a standard pipeline. Based on this comprehensive analysis, we offer the reader insights into potential future research directions. This includes highlighting techniques that show promise due to their effectiveness and identifying areas where further investigation is warranted. To conclude, we highlight the main takeaways from our research and identify any limitations in our analysis.}

\noindent \textbf{Studies distribution by outcome.}
{\color{red}As shown in Figure~\ref{fig:outcome_chart}, segmentation tasks dominate the landscape of FSL studies in medical imaging, accounting for 61\% of the reviewed studies. Classification tasks follow at 32\%, and registration tasks make up the remaining 7\%.}
In the following paragraphs {\color{red}, we delve deeper into how these studies are distributed, considering both the clinical task and the employed meta-learning methods.}

\noindent\textbf{{\color{red} Studies and results distribution by clinical task.}} {\color{red} Figure~\ref{fig:chart_seg_organ} offers an insightful overview of the distribution of clinical tasks related to the segmentation outcome.} Notably, the heart emerges as the most extensively investigated anatomical structure, comprising 34\% of the studies. {\color{red} Following closely behind are the kidney, spleen, and liver, each accounting for 13/\% of the research.} The brain also features significantly, representing 10\% of the studies.
{\color{red} As for classification,} Figure \ref{fig:chart_cls_organ} shows that the lungs take the lead, constituting the primary focus in 36\% of the research. The skin follows closely with 21\%, while the breast and eye account for 15\% and 10\%, respectively.
{\color{red} Lastly, according to Figure \ref{fig:chart_reg_organ}, among registration tasks, the heart emerges as the most commonly investigated,  representing 52\% of the studies focus,} followed by lungs and brain, accounting for 14\% of the studies and the knee for 10\%. Finally, the cervical vertebra and abdomen represent the main application in 5\% of the studies. 

{\color{red} Moving to our statistical analysis of results, Figure \ref{fig:fp_seg_organ} provides insights into the performance of segmentation tasks in terms of the Dice score and IoU across various anatomical structures.} Notably, femur segmentation demonstrates the highest Dice score, although it's worth mentioning that only one study addresses this task, making the result only partially reliable. In contrast, AA  and LV segmentation exhibit consistently good average results across multiple peer studies, achieving Dice scores of 0.89 and 0.88, respectively. The worst-performing segmentation task appears to be the prostate, with a mean Dice score of 0.42 across different structures.
{\color{red} As for the IoU metric, hip, knee, and phalanx segmentation provide the best results. It is worth noting that the present findings are based on a limited pool of studies, potentially impacting their generalizability.} On the other hand, lung segmentation demonstrates a high IoU of 0.91, with a small CI, across several studies, indicating robust and consistent performance. Conversely, prostate segmentation consistently yields lower IoU scores, with a 0.23 IoU for the segmentation of the peripheral zone, indicating room for improvement in this specific task. 
{\color{red}In terms of the classification outcome, even though Figure~\ref{fig:fp_cls_organ} indicates that classifying brain images into different contrast types exhibits the highest accuracy, it's essential to acknowledge that this specific task is considered relatively straightforward. In contrast, the more challenging task of skin lesion classification achieves a high mean accuracy (0.82) across a considerable number of studies. Conversely, liver segmentation demonstrates the lowest average performance, yielding an accuracy of 0.61 across various diseases.
Concerning the AUROC metric, eye image classification achieves the highest performance, scoring 0.96. Conversely, lung segmentation exhibits the lowest average performance (0.69 AUROC across various diseases). However, it is noteworthy that this task involved an ample number of studies.
Finally, it is worth mentioning that a few studies evaluated the classification performance of different lung diseases employing the F1-score and Recall metrics. However, the results in both cases were poor, scoring below 0.5.
Finally, concerning the registration task, Figure~\ref{fig:fp_reg_organ} suggests that FSL achieves its highest registration performance in terms of Dice score for knee bones. However, cautious interpretation is advised due to data from a single study. Conversely, whole heart image registration demonstrates a consistently high mean Dice score (0.88) with a narrow CI, reflecting findings from a larger number of studies. Regarding the ALD and TRE metrics, it's noteworthy that they were exclusively employed for lung image registration.}

\noindent \textbf{{\color{red} Studies and results distributions by meta-learning technique.}} {\color{red} Figure \ref{fig:chart_seg_method} reveals that most studies (55\%) did not utilize any meta-learning method for segmentation. Among those that did, meta-learning approaches (26\%) were the most popular, followed by initialization-based (13\%) and hallucination-based methods (6\%). Shifting the focus on classification tasks, Figure~\ref{fig:chart_cls_method} reveals that 40\% of classification studies did not leverage meta-learning algorithms. However, of those who did (60\%), metric learning-based methods were the most common (40\%), followed by initialization-based methods (20\%). Notably, no studies used hallucination-based methods for classification. Finally, in terms of registration, we found that none of the registration studies employed meta-learning approaches.}

{\color{red}Our statistical analysis of segmentation studies performance (Figure~\ref{fig:fp_seg_method}) reveals interesting findings. Both no-meta-learning methods and hallucination-based methods achieve the highest mean Dice scores (0.84). However, no-meta-learning methods exhibit a wider CI, likely due to their application in a higher number of studies. Metric-learning approaches, the most common meta-learning method in segmentation, show slightly lower performance with a mean Dice score of 0.79 and a larger CI. Concerning the IoU metric, this was only employed for evaluating initialization-based and no-meta-learning methods. In this case, no-meta-learning methods significantly outperform the others, achieving a notably better score with a smaller CI.
Analyzing classification tasks (Figure~\ref{fig:fp_cls_method}), we observe that no-meta-learning studies consistently achieve the highest accuracy (with a narrow CI). Conversely, metric-learning methods exhibit lower performance (mean accuracy of 0.81 and a larger CI). Regarding AUROC, no-meta-learning algorithms again yield the best average score (0.84). Interestingly, both initialization-based and metric-learning methods have a mean AUROC of 0.79, although metric-learning methods have a wider CI. Finally, Figure~\ref{fig:fp_reg_method} depicts the mean performance metrics (ALD, Dice, and TRE) for registration tasks, all utilizing non-meta-learning approaches.}

\noindent \textbf{{\color{red} Additional analyses.}} {\color{red} Our analysis of the amount of labelled training data employed reveals that in segmentation studies, the majority (58\%) incorporate one or more labelled data samples, with a significant portion (40\%) utilizing OSL and only 2\% employing ZSL. A similar pattern emerges in classification studies, where 70\% involve at least one labelled sample, 19\% solely rely on OSL, and 7\% utilize ZSL. However, registration studies diverge, with the primary approach being OSL (50\%). The remaining studies are split between using multiple labelled examples (40\%) and ZSL (17\%).}

In terms of imaging modalities, {\color{red} MRI emerges as the dominant modality across segmentation and registration studies, accounting for 60\% and 83\% of studies, respectively. In contrast, classification studies primarily rely on X-ray imaging, with 41\% of studies utilizing this modality.}

{\color{red} Finally, considering the model's robustness evaluation,} 74\% of the segmentation studies incorporate some form of model robustness evaluation. This typically involves conducting ablation studies and or employing cross-validation techniques. {\color{red} Conversely, only half of the classification and registration studies utilize robustness assessment methods. Ablation studies are the most common approach in these domains.}

\noindent \textbf{{\color{red} Standard pipeline.}} {\color{red} 
As a primary objective, this work aimed to define a standard pipeline reflecting the shared methods employed across all the studies. Building upon the findings from Figure~\ref{fig:standard_pipeline}, in the following paragraphs, we elaborate on the specific arrangement of various learning frameworks within the three pipeline elements: pre-training, training, and data augmentation.}

{\color{red} Among the analyzed studies that utilize pre-training, we identified four main learning paradigms: supervised, unsupervised, self-supervised, and meta-learning. The most common approach, namely supervised learning, involves leveraging pre-trained convolutional backbones on large datasets for transfer learning, as seen in the works of Hansen et al. \cite{Hansen22} and Ali et al. \cite{Ali20}. For unsupervised learning, Chen et al. \cite{Chen21} is the only example. Specifically, they employed an unsupervised cell ranking as a pre-training step for classifying sentinel breast cancer lymph node images. Self-supervised learning is used as a pre-training phase in segmentation studies alone, as exemplified by Tomar et al. \cite{Tomar22}. Here, they pre-trained a style encoder in a self-supervised manner using a volumetric contrastive loss to make it learn clustering similar styled images. The so-trained model is then jointly optimized during the training phase with an appearance and a flow model by minimizing a shared loss.
Finally, we found that meta-learning is also used as a pre-training step, as for Maicas et al. \cite{Maicas19}, who meta-pre-trained their backbone on various tasks before fine-tuning on the task of interest using a fully-supervised approach.

Overall, our analysis revealed that only a few studies employ some pre-training strategy, with supervised learning being the most popular approach. However, while transfer learning offers advantages, several studies highlighted the potential of SSL in outperforming supervised pre-training \cite{Ericsson21,Zhao24}. Ericsson et al. \cite{Ericsson21} demonstrated that leading SSL methods usually surpass supervised pre-training as a source of knowledge transfer. Specifically in the medical domain, Zhao et al. \cite{Zhao24} highlighted that SSL models can exceed baseline models using only 1\% labelled data compared to training from scratch. This efficiency makes SSL particularly valuable in FSL. Additionally, SSL approaches have been shown to learn colour-invariant features \cite{Ericsson21}, a potentially beneficial feature in medical imaging where data is often grayscale. In terms of meta-learning, despite its primary use in downstream task training, its ability to improve generalization can also provide a strong foundation for fine-tuning, as demonstrated by Maicas et al. \cite{Maicas19}. Given these findings, we recommend future FSL research in medical imaging to explore SSL and meta-learning approaches during pre-training for enhanced model capabilities and reduced reliance on labelled data.}

{\color{red} For the main training phase, a wider range of learning frameworks compared to pre-training have been utilized. Indeed, while all four pre-training paradigms (supervised, unsupervised, self-supervised, and meta-learning) are still present, semi-supervised and ZSL approaches are also employed. Meta-learning emerges as the dominant choice, particularly for segmentation and classification. Interestingly, registration studies lean solely on unsupervised or weakly-supervised learning, potentially due to the task nature, which is often addressed by training the model until convergence, mimicking classical registration algorithms. However, outside the FSL context, research by Kanter et al. \cite{Kanter22} and Park et al. \cite{Park22} demonstrates the potential of meta-learning for image registration. This suggests that future research should explore leveraging meta-learning frameworks to enhance the capabilities of registration models in FSL settings, mirroring its success in segmentation and classification tasks.}

{\color{red} Beyond meta-learning, supervised learning remains the second most employed approach for segmentation and classification tasks. This framework utilizes various strategies to enhance model performance and generalizability. In segmentation studies, data augmentation is a popular choice for addressing FSL in a classical supervised way, often involving jointly trained segmentation and registration models where the registration output serves as realistic additional data (e.g., Chan et al. \cite{Chan22}; He et al. \cite{He20}). Other studies, instead, leverage classical data augmentation methods (Guo et al. \cite{Guo22}). Additionally, some works focus on improving the generalization capabilities of the model itself, like Roy et al. \cite{Roy20}, who proposed a two-armed architecture with a "conditioner arm" processing labelled slices to generate task-specific representations to be provided to a "segmenter arm" performing segmentation on query images. 
Similar to segmentation, classification studies often employ data augmentation techniques to achieve good performance with supervised learning frameworks \cite{Huang22, Roychowdhury21}, or rely on robust pre-training with unsupervised or meta-learning approaches, as seen in the works of Chen et al. \cite{Chen21} and Maicas et al. \cite{Maicas19}, respectively.
Unlike segmentation and classification tasks, which primarily rely on supervised learning, only two registration studies employ a supervised approach, i.e., Shi et al. \cite{Shi23} and Xu and Niethammer \cite{Xu19}, who leverage pseudo-labels generated by the segmentation network to train the registration network in a weakly-supervised manner. On the other hand, the most popular learning framework among registration studies involves an unsupervised method where the model is trained until convergence using only the two images to be registered. Interestingly, no segmentation or classification study in our analysis employed unsupervised learning in their training phase. 

The rest of the segmentation and classification studies rely on SSL, ZSL, and semi-supervised learning techniques. These approaches address the data scarcity by leveraging additional information from unlabeled data or by encoding distinguishing properties of objects to be transferred to unseen classes. For instance, Paul et al. \cite{Paul221} leveraged semi-supervised learning and GZSL by proposing a multi-view semantic embedding network that maps a feature embedding extracted from an X-ray image to a semantic signature corresponding to the disease present in the image itself and combines it with a self-training approach to improve diagnosis on unseen classes. Regarding the SSL domain,  we found that, in segmentation studies, the main thread is to couple SSL with a meta-learning framework. An example of this approach is provided by Hansen et al. \cite{Hansen22}, who propose a detection-inspired-network trained self-supervised end-to-end in an episodic manner. On the other hand, in classification, only Mahapatra et al. \cite{Mahapatra22} leveraged SSL, coupling it with supervised learning for GZSL. 

Overall, our analysis highlights a continued reliance on classical supervised learning approaches to address FSL, despite studies demonstrating SSL's potential to achieve supervised-like results using a portion of labelled data \cite{Ericsson21,Zhao24}. Following this, we recommend further exploring the combination of SSL and meta-learning, which provided promising results in the investigated  FSL segmentation studies. Similarly, we think that semi-supervised approaches should be investigated further, given their success on their own in overreaching fully-supervised methods (\cite{Berthelot19,Rasmus15}) and their promising results when used in conjunction with meta-learning (\cite{Ren18,Li19,Boney18}).}

Finally, speaking about data augmentation techniques, {\color{red} our analysis shows that three data-augmentations varieties are employed among FSL for medical imaging studies: classical, registration-based and generative augmentation. Classical data augmentation techniques often regard, among others, geometric image transformations, random erasing, and color jitter. In addition, some studies provided a custom interpretation of classical augmentation, as Guo et al. \cite{Guo22}, who proposed a 3D augmentation of MRI images, which becomes the key to performing kidney segmentation in a supervised way using a limited number of data. Another popular type of data augmentation consists of leveraging registration models jointly trained with segmentation ones and using the output of the registration model as a realistic form of data augmentation to improve segmentation performance \cite{Xu19,Shi23,He20,He22}. Related to registration-based augmentations are also the works of Chan et al. \cite{Chan22} and Shen et al. \cite{Shen20}. In the first one, the authors leverage the classical Daemons registration algorithm to generate additional images and labels to train a segmentation network. In the second one, the authors propose a fluid-based augmentation method which generates anatomically meaningful images via interpolation from the geodesic subspace underlying the provided samples. 

The last data augmentation category we identified among the examined studies is generative-based. This category considers all the studies that propose data generation schemes for data augmentation. For example, Huang et al. \cite{Huang22} proposed the AugPaste framework,  where lesion patches are processed with data augmentation operations and the transformed lesion patches are randomly pasted to the normal samples via MixUp, yielding synthetic anomalous samples. In addition, several segmentation studies combined generative and registration-based approaches by employing image registration w.r.t. an atlas to gain shape deformations and intensity changes between the two images, which are then exploited to synthesize additional images \cite{Ding21,Tomar22,Xu19}.}

\noindent \textbf{{\color{red} Closing remarks.}} {\color{red} Given that the meta-learning framework is the prevailing method for addressing FSL, we devoted a significant portion of our analysis to it. Our investigation revealed variations in the adoption and success rates of different meta-learning techniques. Specifically, methods based on metric-learning have garnered substantial attention, while hallucination-based techniques still require further exploration despite showing promising results. Furthermore, we showed that, on average, non-meta-learning methods outperform meta-learning approaches. Several factors may contribute to this behaviour.
We argue that one of the most likely reasons lies in the inherent nature of meta-learning algorithms. Indeed, meta-learning algorithms are designed to learn a generalized approach to solving tasks by training on diverse tasks. Consequently, there often exists a class mismatch between the tasks seen during training and those encountered during testing. In other words, models are evaluated on tasks involving classes that were never seen during training. Conversely, in classical supervised frameworks, for example, models are specifically trained to excel at a particular task, and the limited availability of data is usually mitigated by leveraging robust augmentation strategies and/or unlabeled data, as in self-supervised and semi-supervised approaches. Therefore, to interpret these results effectively, one must go beyond simple average performance metrics and consider the actual generalizability of the model across different tasks, particularly when additional data are not employed.
}

In terms of the clinical task investigated, our analysis highlights that the heart, abdomen, and lungs have been the primary areas of focus in the examined studies. This is likely due to the availability of well-established benchmark datasets such as CHAOS \cite{Kavur21}, MS-CMRSeg \cite{Zhuang18}, and NIH Chest X-ray \cite{Wang17}. {\color{red} Nevertheless, there is an unexplored potential for researchers to delve into relatively less-explored medical applications, including the prostate and digestive organs. These areas remain critical in clinical practice, and the adoption of FSL techniques could open new avenues for their utilization in the clinical domain.}

{\color{red} Moreover, our observations reveal that certain studies, especially those focusing on classification and registration tasks, may lack comprehensive model investigation analyses. This gap in research practices could result in incomplete and unreliable performance assessments. Additionally, we identified issues pertaining to the reporting of ROB in some studies. Indeed, many of these studies lack clarity in explaining their approach to addressing FSL tasks, even when asserting the utilization of reduced amounts of labelled data.}

In light of our findings, we encourage future researchers in the field to consider the following actions:
\begin{itemize}
    \item {\color{red} Focus the research on the development and improvement of meta-learning methods, as this is the framework that guarantees better generalizability for the model. Specifically, evaluate the combined use of meta-learning and SSL as this showed promising results.} 
    \item {\color{red} Prefer the use of meta-learning, SSL and semi-supervised learning approaches w.r.t. fully-supervised as several studies demonstrated their success in overreaching fully-supervised methods with much less data.}
    \item {\color{red} Among the meta-learning methods, explore the hallucination-based methods in more depth, as they are the least investigated despite being promising.}
    \item {\color{red} Consider the use of meta-learning methods for performing registration tasks, as all the investigated studies at the moment rely on unsupervised and weakly-supervised approaches.}
    \item Expand the scope of medical applications investigated {\color{red} beyond heart, abdomen and lung anatomical structures.}
    \item Prioritize thorough model validation and comprehensive analyses to facilitate fair comparisons and the practical implementation of FSL models in clinical settings.
\end{itemize}

{\color{red} \noindent \textbf{Limitations.} Our study includes some limitations. Primarily, in our research, we focused on multidisciplinary and computer science-related databases, omitting medical databases. This may have affected the comprehensiveness of our literature review. Additionally, we did not encompass FSL papers addressing the image reconstruction task; instead, we concentrated on segmentation, classification, and registration studies.}

\section{Conclusions}
\label{sec:conclusions}
\noindent {\color{red} This work presented a comprehensive analysis of SOTA FSL techniques for the medical imaging domain. We extracted the most relevant information from each study and performed a statistical analysis of the studies' results w.r.t. both the clinical task investigated and the meta-learning algorithm employed. Our work culminated with designing a standard methodological pipeline shared across all the studies. Our findings not only reveal the limitations of existing SOTA methods but also pave the way for future advancements. We advocate for prioritizing approaches that enhance model generalizability and reduce reliance on labelled data, such as self-supervised, semi-supervised, and meta-learning techniques. Additionally, we encourage investigating the potential of hallucination-based methods for FSL in medical imaging. Furthermore, we emphasize the need to expand research across diverse clinical domains as well as employ techniques to asses the robustness and generalizability potential of developed models. These recommendations aim to guide researchers towards achieving significant advancements in this domain and ultimately bridge the gap between FSL and real-world clinical applications.}

\vspace{5mm}

\noindent \textbf{Acknowledgments}\\
This study was funded by the European Union's Horizon 2020 Research and Innovation Program under Grant Agreement No 952159 (ProCAncer-I) and by the Regional Project PAR FAS Tuscany-NAVIGATOR. The funders had no role in the study design, data collection, analysis, interpretation, or manuscript writing.

\printcredits

\bibliographystyle{elsarticle-num}

\bibliography{main}

\end{document}